\documentclass{article}

\usepackage[utf8]{inputenc} 	        
\usepackage{graphicx}      		        
\usepackage{amsfonts}        	        
\usepackage{amsmath,amssymb}            
\usepackage{bm}					        
\usepackage{xcolor}                     
\usepackage{amsthm}                     
\usepackage{algorithm,algpseudocode}    
\usepackage{tabularx}                   
\usepackage{booktabs}                   
\usepackage{caption}
\usepackage{subcaption}                 
\usepackage{url}                        
\usepackage{authblk}                    
\usepackage{placeins}                   

\usepackage[
    style=numeric,
    sorting=none
]{biblatex}                             
\newcommand{\reals}{\mathbb{R}}
\newcommand{\expectation}{\mathrm{E}}
\newcommand{\expectationp}[2][]{\expectation \ifx #1 \undefined \else _{#1} \fi \left[#2\right]}
\newcommand{\variance}{\mathrm{Var}}
\newcommand{\variancep}[2][]{\variance \ifx #1 \undefined \else _{#1} \fi \left[#2\right]}
\newcommand{\dataset}{\mathcal{D}}

\newcommand{\kullb}[2]{D_{\text{KL}} \left( {#1} \, \lVert \, {#2} \right)}
\newcommand{\elbo}{\mathcal{L}}
\newcommand{\condbar}{\,|\,}  
\newcommand{\independence}{\perp\!\!\!\perp}  

\newcommand{\uniformdist}{\mathcal{U}}
\newcommand{\normaldist}{\mathcal{N}}

\title{A deep latent variable model for semi-supervised multi-unit soft sensing in industrial processes}

\author[1, 2, a]{Bjarne Grimstad}
\author[1, 2, b]{Kristian Løvland}
\author[1, c]{Lars S. Imsland}
\author[2, d]{Vidar Gunnerud}
\affil[1]{Norwegian University of Science and Technology, Trondheim, Norway}
\affil[2]{Solution Seeker AS, Oslo, Norway}
\affil[a]{\normalsize \texttt{bjarne.grimstad@ntnu.no} (Corresponding author)}
\affil[b]{\normalsize \texttt{kristian.lovland@ntnu.no}}
\affil[c]{\normalsize \texttt{lars.imsland@ntnu.no}}
\affil[d]{\normalsize \texttt{vidar@solutionseeker.no}}
\date{}

\begin{document}

\maketitle

\begin{abstract}
In many industrial processes, an apparent lack of data limits the development of data-driven soft sensors. There are, however, often opportunities to learn stronger models by being more data-efficient.
To achieve this, one can leverage knowledge about the data from which the soft sensor is learned.
Taking advantage of properties frequently possessed by industrial data, we introduce a deep latent variable model for \textit{semi-supervised} \textit{multi-unit} soft sensing. This hierarchical, generative model is able to jointly model different units, as well as learning from both labeled and unlabeled data.

An empirical study of multi-unit soft sensing is conducted using two datasets: a synthetic dataset of single-phase fluid flow, and a large, real dataset of multi-phase flow in oil and gas wells. We show that by combining semi-supervised and multi-task learning, the proposed model achieves superior results, outperforming current leading methods for this soft sensing problem. We also show that when a model has been trained on a multi-unit dataset, it may be finetuned to previously unseen units using only a handful of data points. In this finetuning procedure, unlabeled data improve soft sensor performance; remarkably, this is true even when no labeled data are available.
\end{abstract}


\section{Introduction}
\label{sec:introduction}
Modern process industries are increasingly concerned with sustainable development to enhance product quality and process efficiency while minimizing adverse environmental impacts \cite{perera_role_2023}. Sustainable development is helped by novel technologies for process monitoring and control, enabled by the ongoing digitalization in the process industries \cite{Yuan2020,Sun2021}. One example is the development of cost-efficient \emph{soft sensors} as an alternative to hardware sensors for process monitoring \cite{Kadlec2009,Jiang2021}. 

A soft sensor is a mathematical model that operates on process measurements ($x$) to make timely inferences about a variable of interest ($y$). Often, $y$ is a key quality or performance indicator that is hard to measure directly and thus is measured occasionally. This may be the case if a measurement requires an expensive or disruptive experiment. Process measurements, $x$, from existing sensors are comparatively cheap and frequent. If $x$ is informative of $y$, a discriminative model $p(y \condbar x)$ can be developed and applied to predict $y$ conditionally on $x$. With this model, soft sensing enables indirect monitoring of $y$ in periods where only $x$ is measured.

In \emph{data-driven} soft sensing, the model is learned from historical data using statistical modeling or machine learning techniques. The attraction to data-driven soft sensing is two-fold: first, it may exploit the increasing amount of data to improve models; second, it promises to reduce costs by simplifying modeling. This is opposed to soft sensing based on mechanistic models or first principles, which may require substantial investments and domain expertise to develop and maintain \cite{Kadlec2009}. In recent years, deep learning has been widely applied to data-driven soft sensing with varying degrees of success \cite{Yuan2020,Sun2021}.

\begin{figure}[bt]
    \centering
    \includegraphics[width=\textwidth]{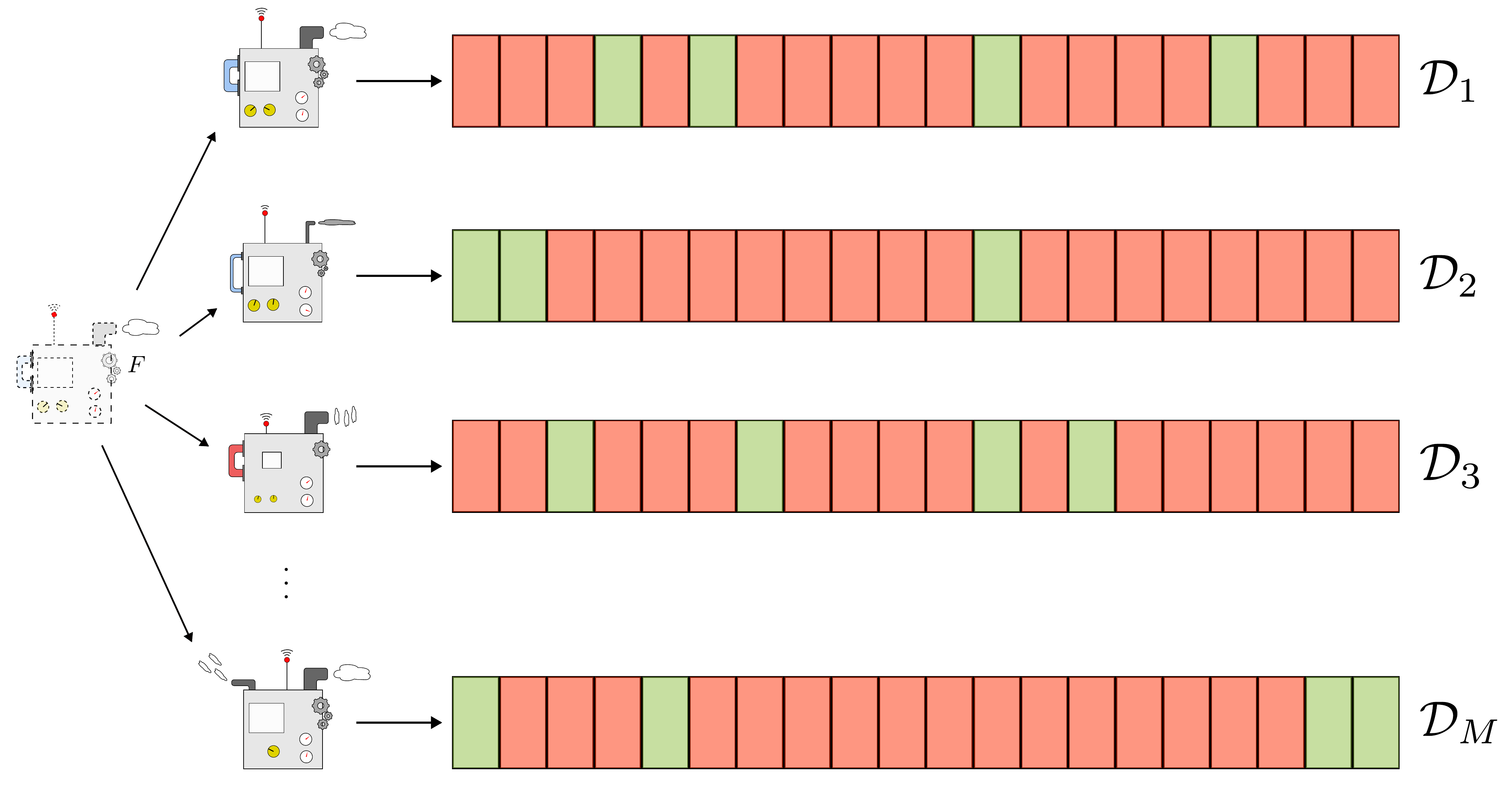}
    \caption{Illustration of multi-unit data generation. The units are similar, and can be thought of as realizations of a common ``prototype" unit. Each unit gives rise to a large amount of unlabeled data (red boxes), and a small amount of labeled data (green boxes).}
    \label{fig:multi-unit-data}
\end{figure}

The main challenge in data-driven soft sensing is arguably to learn a model when the data volume, measurement frequency, variety, or quality is low \cite{Kadlec2009}. The challenge is closely tied to the motivation for implementing soft sensing: to infer a key variable that is measured infrequently. To achieve a good predictive performance in information-poor environments, it is imperative to be \emph{data efficient}. In some circumstances, there is an opportunity to improve data efficiency by modeling on more data. Semi-supervised learning (SSL) and multi-task learning (MTL) are two learning paradigms which have been employed in soft sensing to this end. Consider the multi-unit setting in Figure \ref{fig:multi-unit-data}, where a soft sensor is developed on data collected from multiple process units. In this example, the units may for instance represent similar machines that perform the same operation in multiple production lines. A model learned by SSL and MTL may utilize the whole dataset, learning across units and from unlabeled data. In many cases, one would expect that such a model will be able to exploit the similarity between units as well as the additional unlabeled data and outperform models developed per unit which only use labeled data. 

There is an inherent opportunity for SSL in many soft sensor applications, due to the higher measurement frequency of $x$ (compared to $y$). With an SSL method, the model is learned from both labeled and unlabeled data \cite{chapelle_semi-supervised_2006}. Many SSL methods are based on a generative model of the joint distribution $p(x, y)$. As made clear by the factorization $p(x, y) = p(y \condbar x)p(x)$, generative models are more ambitious in terms of their modeling scope, as they approximate both the conditional distribution $p(y \condbar x)$ and the marginal distribution $p(x)$. SSL methods can provide a powerful regularization mechanism to the model via $p(x)$, given that an a priori dependence between $p(y \condbar x)$ and $p(x)$ is allowed, and that the true data generating system exhibits such a dependence \cite{Seeger2006}. This is not always true; for instance, it was shown in \cite{Scholkopf2012} that semi-supervised learning will not outperform supervised learning if $x$ is a cause of (and not caused by) $y$. In soft sensing problems, there is arguably often not a clear direction of causality, and for dynamical systems and their steady-states, directions of causality may even be interpreted as going both ways \cite{bongers2022causal}. In that case, one can only conclude that semi-supervised learning \textit{may} be advantageous \cite{Scholkopf2012}.

In settings where $p(x)$ does contain information about $p(y \condbar x)$, the advantage of using SSL is largest when the ratio of unlabeled to labeled data is large. However, the additional factor $p(x)$ brings with it additional modeling assumptions, and model misspecification can be harmful to the model performance \cite{VanEngelen2020}. Still, SSL has been successfully used to develop soft sensors for a wide range of processes, including: polymerization process \cite{Xie2020}, debutanizer column process \cite{moreira_de_lima_ensemble_2021}, CTC fermentation process \cite{jin_evolutionary_2021,li_semi-supervised_2022}, ammonia synthesis process \cite{Shen2020,sun_deep_2021,yao_semi-supervised_2023}, biochemical process \cite{Esche2022}, with more references given in \cite{Sun2021}. 

A second opportunity for improved data efficiency is to jointly model multiple process units, as opposed to developing a separate model per unit. If multiple units of the same type are installed in different locations (as illustrated in Figure \ref{fig:multi-unit-data}), then it may be possible to learn across their operational data. The MTL paradigm formalizes learning across tasks; in this example, a task would be to learn a soft sensor model for a single process unit. MTL methods may improve data efficiency by sharing statistical strength between learning tasks. In cases where there is little data per task, but many tasks, MTL may significantly outperform single-task learning methods \cite{Zhang2021}. However, because MTL applied to multi-unit data requires data from multiple units, it has been used less than SSL in soft sensing applications. A survey on transferability in soft sensing is given in \cite{Curreri2021}, and more recently in \cite{grimstad2023multi}. Transfer learning and MTL have recently been used in the development of soft sensors for different processes, including: polyethylene process \cite{Liu2019}, virtual flow metering \cite{Sandnes2021}, solid waste incineration process \cite{qiao_multitask_2023}, and a sulfur recovery unit process \cite{huang_modeling_2023}. 

This paper shows how to leverage the two learning paradigms of MTL and SSL to obtain a data-efficient method for data-driven soft sensing. We model the data using a deep latent variable model (DLVM), a type of generative model for which conditional probability distributions are represented by deep neural networks \cite{Kingma2019}. The generic model relies on two hierarchical levels of latent variables, $c$ and $z$, to explain the observations $(x, y)$. Building on the framework of the variational auto-encoder \cite{Kingma2014a}, we show how to make inferences for latent variables and estimate the model parameters from data. The main use case of our model and the focus of this paper is online prediction, for which accurate and timely inferences of $y$ are required. In addition to the use we consider here, the generative model has additional applications relevant to soft sensing, which are not admitted by discriminative models. First, it enables data generation, which can be used in what-if scenario analyses. Second, it can be used to impute missing data. A third use is sensor surveillance, where data deemed improbable by the model may indicate sensor faults.


Our method is based on the variational autoencoder (VAE), which was first adapted to semi-supervised learning in \cite{Kingma2014b}. Several variants of semi-supervised VAE have later been developed \cite{Yang2021}. Most of these are specialized to semi-supervised classification. There are some notable works on the combination of SSL and MTL for both classification and regression problems \cite{Liu2007,Zhang2009,Zhuang2015}. Of these, the SAML method for classification, presented in \cite{Zhuang2015}, is perhaps closest to our method. Within data-driven soft sensing, the combination of SSL and transfer learning across operating conditions of a single process unit in \cite{chai_deep_2022} is close in spirit to our method. Our work is different since it combines SSL with MTL in a multi-unit soft sensor setting. To the authors` knowledge, no previous work have explored this.

\subsection{Organization of the paper}
The paper is structured as follows. Section \ref{sec:problem-statement} presents the problem statement. The model is presented in Section \ref{sec:model} and the learning method in Section \ref{sec:learning-method}. A comparison to related methods is given in Section \ref{sec:related-methods}. An empirical study of the method is given in Section \ref{sec:empirical-study} and the results are discussed in Section \ref{sec:discussion}. Finally, some concluding remarks are given in Section \ref{sec:conclusion}.

\subsection{Notation}
Sets of variables are compactly represented by bold symbols, e.g. $\bm{c} = \{c_i\}_i$. For double-indexed variables we write $\bm{x} = \{x_{ij}\}_{ij}$ and $\bm{x}_i = \{x_{ij}\}_{j}$, which permits us to also write $\bm{x} = \{ \bm{x}_i \}_i$. When the indices are irrelevant to the discussion, we may write $x$ instead of $x_{ij}$.

For a positive integer $K$, we denote the $K$-vector of zeros by $0_K$ and the $K\times K$ identity matrix by $I_K$. For a $K$-vector $\sigma = (\sigma_1, \ldots, \sigma_K)$, $\log \sigma$ denotes the element-wise logarithm, and $\text{diag}(\sigma^{2})$ denotes the diagonal $(K\times K)$-matrix with diagonal elements $(\sigma_{1}^{2}, \ldots, \sigma_{K}^{2})$. 

The normal distribution is denoted by $\normaldist(\mu, \Sigma)$, where $\mu$ and $\Sigma$ is the mean and covariance matrix, respectively. 
For a random variable $X \sim p(x)$, we denote the expectation operator by $\expectation[X]$. We will sometimes write $\expectationp[X]{X}$ or $\expectationp[p]{X}$ to indicate that the expectation is of a specific random variable or probability distribution.

\section{Problem statement}
\label{sec:problem-statement}
Consider a set of $M > 1$ distinct, but related \emph{units} indexed by $i \in \{1, \ldots, M\}$. The units may be a set of objects, e.g. pumps, valves, or solar panels, or a set of complex systems, such as distillation columns, oil wells, or robots. The units are assumed to be related so that it is reasonable to expect, a priori to seeing data, that transfer learning between units is beneficial. 

We assume that the same explanatory variables $x \in \reals^{D_x}$ and target variables $y \in \reals^{D_y}$ can be considered for all units. In soft sensor applications, $x$ typically represents cheap and abundant measurements from which we wish to infer a target variable $y$ which is expensive or difficult to measure. For each unit $i$, we have at our disposal a set of $N_i^l$ \emph{labeled} data $\dataset_{i}^{l} = \{(x_{ij}^{l}, y_{ij}^{l})\}_{j}$ and a set of $N_i^u$ \emph{unlabeled} data $\dataset_{i}^{u} = \{x_{ij}^{u}\}_{j}$. We collect the $N_i = N_i^l + N_i^u$ data points of unit $i$ in the dataset $\dataset_i = \dataset_{i}^{l} \cup \dataset_{i}^{u}$. We assume that $\dataset_i$ consists of independent and identically distributed (i.i.d.) observations drawn from a probability distribution $p_i$ over $\reals^{D_x \times D_y}$. 

Our goal is to efficiently learn the behavior of the units, as captured by $p_1, \ldots, p_M$, from the data collection $\dataset = \{\dataset_1, \ldots, \dataset_M\}$. In particular, we wish to exploit that some structures or patterns are common among units. Furthermore, to fully utilize the information in $\dataset$, we wish to learn from unlabeled data, which are overrepresented in most soft sensor applications. 

Invariably, distinct units will differ in various ways, e.g., they may be of different design or construction, they may be in different condition, or they may operate under different conditions. The properties or conditions that make units distinct are referred to as the \emph{context} of a unit. An observed context may be included in the dataset as an explanatory variable. An unobserved context must be treated as a latent variable to be inferred from data and it is only possible to make such inferences by studying data from multiple, related units. We assign the letter $c$ to latent context variables.

We will assume that there is some underlying process $p$ at population level, such that $p_i = p(\cdot \condbar c_i)$, where $c_i \in \reals^{K}$ is a latent variable drawn from $p(c)$. The variable $c_i$ represents unobserved properties or context of unit $i$. We can now restate our goal as learning from $\dataset$ a generative model $\hat{p}(x,y)$ which approximates $p(x,y) = \int p(x, y \condbar c) p(c)$.

\section{A generative model for multi-unit soft sensing}
\label{sec:model}
\begin{figure}[bt]
\centering
\includegraphics[width=0.4\linewidth]{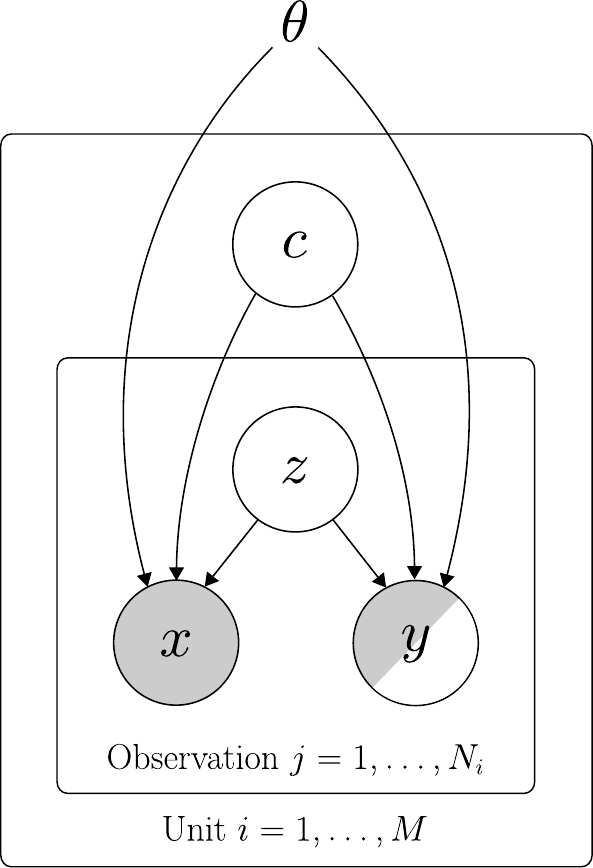}
\caption{A generative model for multi-unit soft sensing. Random variables are encircled. A grey (white) circle indicates that the variable is observed (latent). The nested plates (rectangles) group variables at different levels.}
\label{fig:gen-model}
\end{figure}

We consider a hierarchical model for generating observations of a population of units. Observations are split semantically so that $x$ labels cheap and frequent observations, and $y$ labels expensive and occasional observations. The model has three levels of latent variables that explain the observations. At observation level, $z$ represents a latent state which is (partly) observed by $x$ and $y$. At unit level, the latent variable $c$ represents the context or specificity of a unit. At the universal top-level, $\theta$ represents properties which are shared by all units and observations. The hierarchical structure allows the model to capture variations at the level of units and observations. A graphical illustration is given in Figure \ref{fig:gen-model}. In the figure, $\theta$ is treated as a parameter.

The model assumes the following generative process for a collection $\dataset$ of observations from $M$ units. For each unit $i = 1, \ldots, M$:
\begin{equation}
\begin{aligned}
    c_{i} &\sim p(c) = \normaldist(0_K, I_K), & \\
    z_{ij} &\sim p(z) = \normaldist(0_D, I_D), & j=1,\ldots,N_{i}, \\
    x_{ij} \condbar z_{ij}, c_{i} &\sim p_{\theta}(x \condbar z_{ij}, c_{i}), & j=1,\ldots,N_{i}, \\
    y_{ij} \condbar z_{ij}, c_{i} &\sim p_{\theta}(y \condbar z_{ij}, c_{i}), & j=1,\ldots,N_{i}.
\end{aligned}
\label{eq:generative-model}
\end{equation}

At the unit level, latent context variables $c_i \in \reals^K$ are generated from a common prior distribution $p(c) = \normaldist(0_K, I_K)$. At the level of observation, latent variables $z_{ij} \in \reals^D$ are generated from a common prior $p(z) = \normaldist(0_D, I_D)$. The standard normal priors on the latent variables allow for easy sampling and reparameterization, and is commonly used in variational autoencoders \cite{Kingma2019}. 

Observations $x$ and $y$ are generated conditionally independent given the latent variables $z$ and $c$. That is, $p_{\theta}(x, y \condbar z, c) = p_{\theta}(x \condbar z, c) p_{\theta}(y \condbar z, c)$. For the continuous variables $x$ and $y$, an appropriate likelihood function may be the multivariate normal
\begin{align*}
    p_{\theta}(x, y \condbar z, c) = \normaldist(\mu_{\theta}(z, c), \Sigma_{\theta}(z, c)),
\end{align*}
where $\mu_{\theta}(z, c)$ and $\Sigma_{\theta}(z, c)$ are neural networks with parameters $\theta$. The resulting model is a type of deep latent variable model and provides an expressive family of conditional distributions for modeling $x, y \condbar z, c$. 

Several simplifying assumptions are made in the above model: 1) the dimensions of the latent variables, $K$ and $D$, are fixed and considered to be choices of design; 2) $\theta$ is treated as a parameter vector to be estimated, not as a latent variable with a prior $p(\theta)$; 3) no hyper-priors are put on the parameters of $p(z)$ and $p(c)$. In a fuller Bayesian approach, it would be natural to remove assumption 2 and 3, and include a hyper-prior also for $p(\theta)$.

\subsection{Missing target values and grouping of variables}
As indicated in Figure \ref{fig:gen-model} by the partly shaded circle for $y$, targets may be observed or missing. For unit $i$, we denote an observed target by $y_{ij}^l$ and a missing target by $y_{ij}^u$. Corresponding observations of $x$ are denoted by $x_{ij}^l$ and $x_{ij}^u$. We group these variables in the sets $\bm x_i^l$, $\bm y_i^l$, $\bm x_i^u$, and $\bm y_i^u$. Note that $\bm x_i = \bm x_i^l \cup \bm x_i^u$ and $\bm y_i = \bm y_i^l \cup \bm y_i^u$.

We group observed variables for unit $i$ in $\dataset_i = \{\dataset_i^l, \dataset_i^u\}$, where $\dataset_i^l = \{(x_{ij}^l, y_{ij}^l)\}_j$ is the labeled data and $\dataset_i^u = \{x_{ij}^u\}_j$ is the unlabeled data. The remaining variables for unit $i$ are the latent variables which we group in $\mathcal{U}_i = \{\bm y_i^u, \bm z_i, c_i\}$. 

When considering all units, the subscript $i$ is dropped and we write $\dataset$ for all observed variables and $\mathcal{U}$ for all latent variables. Any one variable in $\bm x$, $\bm y$, $\bm z$, or $\bm c$ will be either in $\dataset$ or $\mathcal{U}$.

\subsection{Joint distribution and marginal likelihood}
The joint distribution of all variables in \eqref{eq:generative-model} factorizes as follows
\begin{equation}
\begin{aligned}
    p_{\theta}(\dataset, \mathcal{U}) &= \prod\limits_{i=1}^{M} p(c_i) \prod\limits_{j=1}^{N_{i}} p_{\theta}(x_{ij} \condbar z_{ij}, c_{i}) p_{\theta}(y_{ij} \condbar z_{ij}, c_{i}) p(z_{ij}).
    \label{eq:joint-pdf}
\end{aligned}    
\end{equation}

The \emph{marginal likelihood} of $\dataset$ is obtained by marginalizing out all latent variables $\mathcal{U}$ in \eqref{eq:joint-pdf}:
\begin{align}
    p_{\theta}(\dataset) = \prod\limits_{i=1}^{M} \left( \int \prod\limits_{(x^l, y^l) \in \dataset_i^l} p_{\theta}(x^l, y^l \condbar c) \prod\limits_{x^u \in \dataset_i^u} p_{\theta}(x^u \condbar c) p(c) dc \right),
    \label{eq:marginal-likelihood}
\end{align}
where 
\begin{align*}
    p_{\theta}(x, y \condbar c) &= \int p_{\theta}(x \condbar z, c) p_{\theta}(y \condbar z, c) p(z) dz, \\
    p_{\theta}(x \condbar c) &= \int p_{\theta}(x, y \condbar c) dy = \int p_{\theta}(x \condbar z, c) p(z) dz.
\end{align*}

The calculation in \eqref{eq:marginal-likelihood} is intractable due to the integration over high-dimensional and complex conditional distributions that are parameterized by deep neural networks. Maximum likelihood estimation of $\theta$ is thus not possible. In the following section, we derive an approximate inference procedure to enable the estimation of $\theta$.

\section{Learning method}
\label{sec:learning-method}
Application of Bayes' theorem to the model in \eqref{eq:joint-pdf} results in an alternative expression for the marginal log-likelihood:
\begin{equation}
    \log p_{\theta}(\dataset) = \log p_{\theta}(\dataset, \mathcal{U}) - \log p_{\theta}(\mathcal{U} \condbar \dataset).
    \label{eq:posterior}
\end{equation}

The above expression includes the posterior distribution of the latent variables, $p_{\theta}(\mathcal{U} \condbar \dataset)$. Since the model in \eqref{eq:joint-pdf} is tractable by design, it follows from the intractability of the marginal distribution in \eqref{eq:marginal-likelihood} that the posterior must be intractable. Exact inference of the latent variables is thus unachievable. This is a general issue with DLVMs and motivates the use of approximate inference methods. 

In \emph{variational inference}, the posterior is approximated by an \emph{inference model} $q_{\phi}(\mathcal{U} \condbar \dataset)$ with variational parameters $\phi$. The inference model is used to derive the following lower bound on the marginal log-likelihood:

\begin{align}
    \log p_{\theta}(\dataset) &\geq \elbo_{\theta, \phi}(\dataset) := \expectationp[q_{\phi}]{\log p_{\theta}(\dataset, \mathcal{U}) - \log q_{\phi}(\mathcal{U} \condbar \dataset)},
    \label{eq:elbo}
\end{align}
where the expectation is over values of latent variables $\mathcal{U}$ drawn from $q_{\phi}$. A derivation and discussion of this bound, which often is referred to as the evidence lower bound (ELBO), can be found in \cite{Blei2003}.


Maximization of the ELBO with respect to the parameters $(\theta, \phi)$ achieves two goals: i) it maximizes the marginal likelihood, and ii) it minimizes the KL divergence between the approximation $q_{\phi}(\mathcal{U} \condbar \dataset)$ and the posterior $p_{\theta}(\mathcal{U} \condbar \dataset)$. Another advantage with this strategy is that it permits the use of stochastic gradient descent methods which can scale the learning process to large datasets.

The next sections are spent on the derivation of the inference model and optimization method. Implementation details are provided at the end of the chapter.


\subsection{Inference model}
Our choice of inference model removes some of the complicating dependencies in the posterior. To motivate our simplifications, we consider the following factorization of the posterior distribution:
\begin{equation}
    p_{\theta}(\mathcal{U} \condbar \dataset) = p_{\theta}(\bm{z} \condbar \dataset, \bm{y}^u, \bm{c}) p_{\theta}(\bm{y}^u \condbar \dataset, \bm{c}) p_{\theta}(\bm{c} \condbar \dataset).
    \label{eq:posterior-factorization}
\end{equation}

The first factor in \eqref{eq:posterior-factorization} can be written as:
\begin{align}
    p_{\theta}(\bm{z} \condbar \dataset, \bm{y}^u, \bm{c}) &= \prod\limits_{i=1}^{M} \prod\limits_{j=1}^{N_i} p_{\theta}(z_{ij} \condbar x_{ij}, y_{ij}, c_i).
    \label{eq:posterior-factor-z}
\end{align}
From this expression, we see that the latent variable $z_{ij}$ is inferred from $x_{ij}$, $y_{ij}$ and $c_i$. Evidently, inference of $z_{ij}$ requires access to $y_{ij}$, which may be observed ($y_{ij} \in \bm{y}^l$) or inferred ($y_{ij} \in \bm{y}^u$). The posterior factors in \eqref{eq:posterior-factor-z} are approximated by a variational distribution $q_{\phi_z}(z \condbar x, y, c)$ with parameters $\phi_z$. 

The second factor in \eqref{eq:posterior-factorization} can be written as:
\begin{equation}
    p_{\theta}(\bm{y}^u \condbar \dataset, \bm{c}) = \prod\limits_{i=1}^{M} \prod\limits_{j=1}^{N_i^u} p_{\theta}(y_{ij}^{u} \condbar x_{ij}^{u}, c_i).
    \label{eq:posterior-factor-y}
\end{equation}
A variational distribution $q_{\phi_y}(y \condbar x, c)$ with parameters $\phi_y$ is used to approximate the posterior factors in \eqref{eq:posterior-factor-y}. Notice the resemblance to a discriminative model, where $y$ is conditioned on an input $x$ and context $c$.


The third factor in \eqref{eq:posterior-factorization} can be written as:
\begin{equation}
    p_{\theta}(\bm{c} \condbar \dataset) = \prod\limits_{i=1}^{M} p_{\theta}(c_i \condbar \dataset_i).
    \label{eq:posterior-factor-c}
\end{equation}
The latent variable $c_i$ is inferred from all observations in $\dataset_i$ and can be thought of as a vector of summary statistics of the dataset. Conditioning the latent variable on the full dataset can be challenging for large datasets and it requires a model that can handle different dataset cardinalities. The inference is simplified by allowing an unconditional approximation of the form $q_{\phi_{ci}}(c_i)$, where $\phi_{ci}$ are parameters related to variable $c_i$. An argument for this simplification is that the latent variable is likely to change little when new data points are added to the dataset; i.e., as evidence is accumulated for the context variable $c_i$, the variance of $p_{\theta}(c_i \condbar \dataset_i)$ is expected to shrink.

\begin{figure}[bt]
\centering
\includegraphics[width=0.5\linewidth]{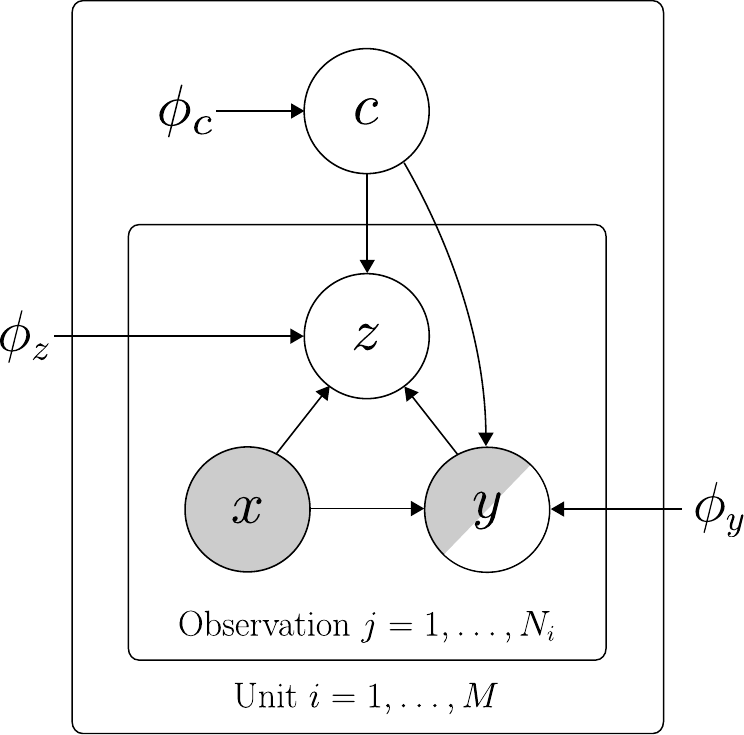}
\caption{A graphical representation of the inference model. Arrows indicate which parameters $(\phi_y, \phi_z, \phi_c)$ and which variables (encircled) that are used in the inference of the latent variables $(y^u, z, c)$.}
\label{fig:inf-model}
\end{figure}

To summarize, we have motivated the following inference model to approximate \eqref{eq:posterior-factorization}:
\begin{align}
    q_{\phi}(\mathcal{U} \condbar \dataset) &= \prod\limits_{i=1}^{M} q_{\phi_{ci}}(c_i) \prod\limits_{j=1}^{N_i^u} q_{\phi_y}(y_{ij}^{u} \condbar x_{ij}^{u}, c_i) \prod\limits_{j=1}^{N_i} q_{\phi_z}(z_{ij} \condbar x_{ij}, y_{ij}, c_i),
    \label{eq:inference-model}
\end{align}
where we have denoted the variational parameters by $\phi = \{\phi_z, \phi_y, \phi_c\}$ and $\phi_c = \{\phi_{c1}, \ldots, \phi_{cM}\}$. The inference model is shown graphically in Figure \ref{fig:inf-model}.

In the following, the variational distributions in \eqref{eq:inference-model} are chosen to be mean-field normal:
\begin{align}
    q_{\phi_{ci}}(c_i) &= \normaldist(m_{ci},  \text{diag}(s_{ci}^{2})), \text{ for } i = 1,\ldots, M,\\
    q_{\phi_y}(y \condbar x, c) &= \normaldist(m_{y}, \text{diag}(s_{y}^{2})), \\
    q_{\phi_z}(z \condbar x, y, c) &= \normaldist(m_{z}, \text{diag}(s_{z}^{2})),
\end{align}
where we have used the symbols $m$ and $s$ for the mean and standard deviation, respectively. The conditional distributions of $y$ and $z$ are computed as $(m_{y}, \log s_{y}) = g_{\phi_y}(x, c)$ and $(m_{z}, \log s_{z}) = g_{\phi_z}(x, y, c)$, where $g_{\phi_y}$ and $g_{\phi_z}$ are neural networks. The \emph{inference networks} amortize the cost of inference by utilizing a set of global variational parameters ($\phi_y$ and $\phi_z$), instead of computing variational parameters per data point. In variational autoencoders, an inference network is usually called an \emph{encoder}.

The parameters of the unconditional variational distribution for $c_i$ is simply computed as $(m_{ci}, \log s_{ci}) = \phi_{ci}$. Thus, the distribution of the context variable $c_i \in \reals^K$ has $2K$ parameters for each task $i$.

With the inference model fully described, a leaner notation is afforded and subsequently we refer to the variational distributions by $q_{\phi}(\cdot)$, where the arguments indicate included factors and parameters. For example, we write $q_{\phi}(y, z \condbar x, c) = q_{\phi_z}(z \condbar x, y, c) q_{\phi_y}(y \condbar x, c)$.

\subsection{Objective function}
With the inference model in \eqref{eq:inference-model}, the ELBO in \eqref{eq:elbo} becomes:
\begin{align}
    \elbo_{\theta, \phi}(\dataset) &= \sum_{i=1}^M \elbo_{\theta, \phi}(\dataset_i),
    \label{eq:model-elbo}
\end{align}
where the terms related to unit $i$ are collected in
\begin{equation}
\begin{aligned}
    \elbo_{\theta, \phi}(\dataset_i) &:= \sum_{x^u \in \dataset_i^u} \expectationp[q_{\phi}(y, z, c_i \condbar x^u)]{\log \frac{p_{\theta}(x^u, y, z \condbar c_i)}{q_{\phi}(y, z \condbar x^u, c_i)}} \\
    &\quad + \sum_{x^l, y^l \in \dataset_i^l} \expectationp[q_{\phi}(z, c_i \condbar x^l, y^l)]{\log \frac{p_{\theta}(x^l, y^l, z \condbar c_i)}{q_{\phi}(z \condbar x^l, y^l, c_i)}} \\
    &\quad - \kullb{q_{\phi}(c_i)}{p_{\theta}(c_i)}.
\end{aligned}
\label{eq:model-elbo-terms}
\end{equation}

Notice that the expectations in $\elbo_{\theta, \phi}(\dataset_i)$ are over the context variable $c_i$ of unit $i$. The final term is the Kullback-Leibler divergence of the variational distribution $q_{\phi}(c_i)$, which can be computed analytically when both $q_{\phi}(c_i)$ and $p_{\theta}(c_i)$ are normal distributions.



\subsection{Optimization method}
\label{sec:optimization-method}
We wish to optimize the ELBO in \eqref{eq:model-elbo} using a gradient-based method. This requires us to compute
\begin{align}
    \nabla_{\theta, \phi} \elbo_{\theta, \phi}(\dataset) &= \nabla_{\theta, \phi} \expectationp[q_{\phi}]{\log p_{\theta}(\dataset, \mathcal{U}) - \log q_{\phi}(\mathcal{U} \condbar \dataset)},
    \label{eq:elbo-gradient}
\end{align}
where the operator $\nabla_{\theta, \phi}$ produces the gradient with respect to the parameters $\theta$ and $\phi$. The gradient computation in \eqref{eq:elbo-gradient} is problematic since the expectation on the right-hand side depends on $\phi$. A common trick to circumvent this problem is to reparameterize the latent variables, here $\bm{y}^u$, $\bm{z}$, and $\bm{c}$. 
Utilizing the reparameterization trick, a one-sample Monte-Carlo approximation of the ELBO is derived:
\begin{align}
    \elbo_{\theta, \phi}(\dataset) \simeq \tilde{\elbo}_{\theta, \phi}(\dataset) = \sum\limits_{i=1}^{M} \tilde{\elbo}_{\theta, \phi}(\dataset_i),
    \label{eq:sgvb-estimator}
\end{align}
where
\begin{equation}
\begin{aligned}
    \tilde{\elbo}_{\theta, \phi}(\dataset_i) &:= \sum_{x^u \in \dataset_i^u} \log \frac{p_{\theta}(x^u, \tilde{y}, \tilde{z} \condbar \tilde{c}_i)}{q_{\phi}(\tilde{y}, \tilde{z} \condbar x^u, \tilde{c}_i)} \\
    &\quad + \sum_{x^l, y^l \in \dataset_i^l} \log \frac{p_{\theta}(x^l, y^l, \tilde{z} \condbar \tilde{c}_i)}{q_{\phi}(\tilde{z} \condbar x^l, y^l, \tilde{c}_i)} \\
    &\quad - \kullb{q_{\phi}(c_i)}{p_{\theta}(c_i)}.
\end{aligned}
\label{eq:sgvb-estimator-terms}
\end{equation}

In \eqref{eq:sgvb-estimator-terms}, the context variable is sampled once (per unit) by computing $\tilde{c}_i = m_{ci} + s_{ci} \odot \varepsilon_{c}$, where $\varepsilon_{c} \sim \normaldist(0_K, I_K)$. The latent variables $z$ and $y^u$ are sampled per data point using the inference networks: 
\begin{equation}
\begin{aligned}
    (m_{y}, \log s_{y}) &= g_{\phi_y}(x, \tilde{c}_i), \\
    \varepsilon_{y} &\sim \normaldist(0_{D_y}, I_{D_y}), \\
    \tilde{y} &= m_{y} + s_{y} \odot \varepsilon_{y}, \\
    (m_{z}, \log s_{z}) &= g_{\phi_z}(x, \tilde{y}, \tilde{c}_i), \\
    \varepsilon_{z} &\sim \normaldist(0_D, I_D), \\
    \tilde{z} &= m_{z} + s_{z} \odot \varepsilon_{z}.
\end{aligned}
\end{equation}
Note that in the second summation in \eqref{eq:sgvb-estimator-terms} both $x$ and $y$ are observed, and $z$ is sampled by first evaluating $(m_{z}, \log s_{z}) = g_{\phi_z}(x, y, \tilde{c}_i)$.

\begin{figure}[bt]
    \centering
    \includegraphics[width=\textwidth]{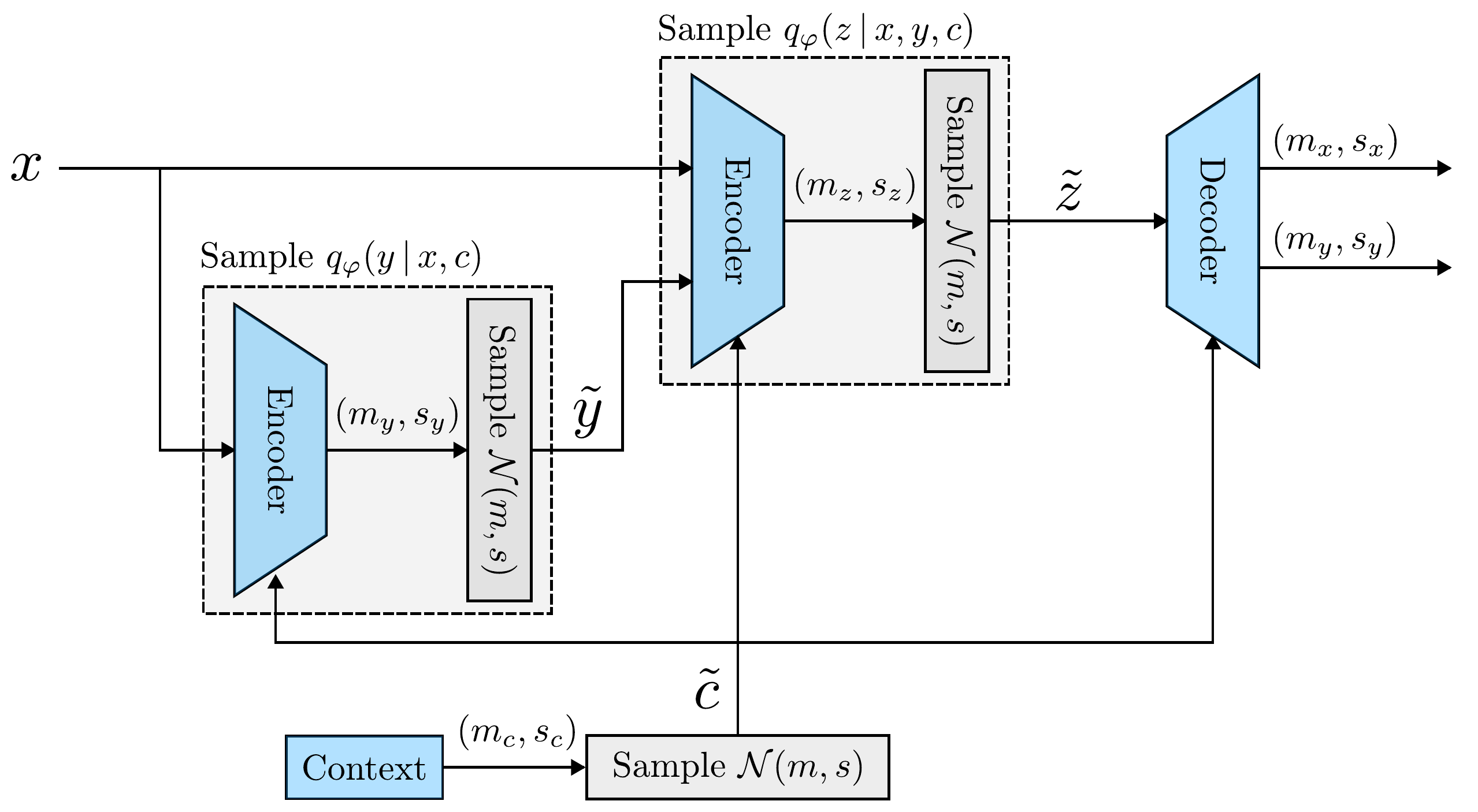}
    \caption{Illustration of model architecture when $y$ is unobserved. When $y$ is observed, the encoder $q_{\varphi}(y \condbar x, c)$ is unused, and $y$ is fed straight into the encoder $q_{\varphi_z}(z \condbar x, y, c)$. In either case, the output is an approximation of $p_{\theta}(x, y)$.}
    \label{fig:model-architecture}
\end{figure}

The gradient of the estimator in \eqref{eq:sgvb-estimator} is known as the Stochastic Gradient Variational Bayes (SGVB) estimator \cite{Kingma2014a}. It is an unbiased estimator of the ELBO gradient, $\nabla_{\theta, \phi} \elbo_{\theta, \phi}(\dataset)$, and can be computed efficiently using back-propagation due to the reparameterization. Pseudocode for the estimator is given in Algorithm \ref{alg:sgvb}, and information flow through the model is illustrated in Figure \ref{fig:model-architecture}.

\begin{algorithm}[bt]
\caption{SGVB estimator for the proposed model}
\label{alg:sgvb}
\begin{algorithmic}[1]
\Require data $\dataset$, generative model $p_{\theta}$, and inference model $q_{\phi}$. 
\State $\tilde{\elbo} \gets 0$ 
\For{$i=1,\ldots,M$} 
\State $\varepsilon_{c} \sim \normaldist(0_K, I_K)$
\State $\tilde{c}_i \gets m_{ci} + s_{ci} \odot \varepsilon_{c}$ 
\For{$x^u \in \dataset_i^u$} \Comment{Unlabeled data}
\State $(m_{y}, \log s_{y}) \gets g_{\phi_y}(x^u, \tilde{c}_i)$
\State $\tilde{y} \gets m_{y} + s_{y} \odot \varepsilon_{y}$
\State $(m_{z}, \log s_{z}) \gets g_{\phi_z}(x^u, \tilde{y}, \tilde{c}_i)$
\State $\tilde{z} \gets m_{z} + s_{z} \odot \varepsilon_{z}$
\State $\tilde{\elbo} \gets \tilde{\elbo} + \log \frac{p_{\theta}(x^u, \tilde{y}, \tilde{z} \condbar \tilde{c}_i)}{q_{\phi}(\tilde{y}, \tilde{z} \condbar x^u, \tilde{c}_i)}$
\EndFor
\For{$(x^l, y^l) \in \dataset_i^l$} \Comment{Labeled data}
\State $(m_{z}, \log s_{z}) \gets g_{\phi_z}(x^l, y^l, \tilde{c}_i)$
\State $\tilde{z} \gets m_{z} + s_{z} \odot \varepsilon_{z}$
\State $\tilde{\elbo} \gets \tilde{\elbo} + \log \frac{p_{\theta}(x^l, y^l, \tilde{z} \condbar \tilde{c}_i)}{q_{\phi}(\tilde{z} \condbar x^l, y^l, \tilde{c}_i)}$
\EndFor
\State $\tilde{\elbo} \gets \tilde{\elbo} - \kullb{q_{\phi}(c_i)}{p_{\theta}(c_i)}$ \Comment{Computed analytically}
\EndFor
\State Compute $\nabla_{\theta, \phi} \tilde{\elbo}$ using back-propagation \\
\Return $\nabla_{\theta, \phi} \tilde{\elbo}$
\end{algorithmic}
\end{algorithm}

\subsection{Augmenting the objective function to emphasize inference of $y$}
As can be seen by inspecting \eqref{eq:model-elbo-terms}, the inference model $q_{\phi_y}(y \condbar x, c)$ only enters in terms with unlabeled data. If the inference model is to be used for prediction, it is unfortunate that it will not be trained directly on labeled data. This issue, which persists for many different generative models developed for semi-supervised learning \cite{Kingma2014b,Siddharth2017,Gordon2020}, is commonly resolved by augmenting the ELBO with a labeled data log-likelihood term for the inference model. For our model, this corresponds to adding the following term to the ELBO:
\begin{equation}
    J_{\theta, \phi}(\dataset) = \frac{1}{N} \elbo_{\theta, \phi}(\dataset) + \alpha \frac{1}{N^l} \sum\limits_{i=1}^{M} \sum_{x^l, y^l \in \dataset_i^l} \expectationp[q_{\phi}(c_i)]{\log q_{\phi}(y^l \condbar x^l, c_i)}.
    \label{eq:augmented-model-elbo}
\end{equation}
Above, $\alpha$ was introduced as a weighting factor to balance the influence of labeled data relative to unlabeled data on the objective function. Notice that the two terms are scaled by the respective number of data points used in their computation, $N$ and $N^l$.

The stochastic gradient variational Bayes method in Section \ref{sec:optimization-method} also applies to the augmented objective in \eqref{eq:augmented-model-elbo}. We only need to form a Monte-Carlo approximation of the log-likelihood term using the same reparameterization trick to draw context variables, $\tilde{c}_i$. The gradient of both terms can then be computed using back-propagation.

The augmented objective function gives us a new perspective on the ELBO. If we consider our main objective to be the learning of a discriminative model, $q_{\phi}(y | x, c)$, so that we can make inferences of $y$ in online prediction, then the ELBO term in \eqref{eq:augmented-model-elbo} can be viewed as nothing more than an advanced regularization term. With this interpretation, the z-encoder and decoder, which are used to form the ELBO, can be discarded after training as their only purpose is to regularize the y-encoder (i.e. the discriminative model) during training.

\section{Comparison to related generative models}
\label{sec:related-methods}
In this section, we compare the model proposed in Section \ref{sec:model} to some closely related generative models found in the literature. The discussed models extend the variational autoencoder (VAE), originally presented for unsupervised learning, to new modeling paradigms. To simplify the comparisons, we focus more on the architectures of the generative models and less on the related inference methods. We also briefly consider the connection between our proposed model and other hierarchical models for multi-task learning. 

\begin{figure}[bt]
     \centering
     \begin{subfigure}[b]{0.3\textwidth}
         \centering
         \includegraphics[width=\textwidth]{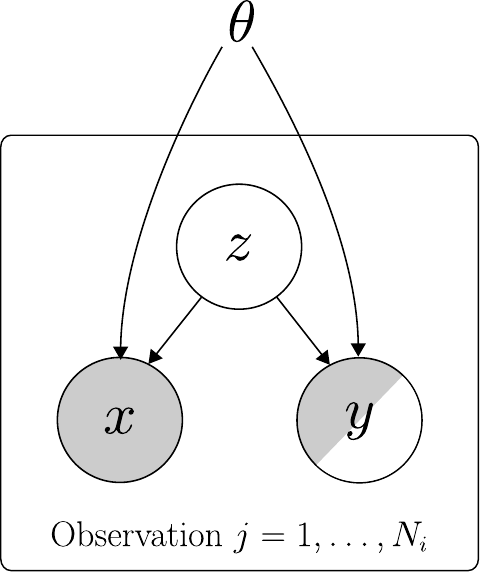}
         \caption{SSL-VAE}
         \label{fig:ssl-vae}
     \end{subfigure}
     \hfill
     \begin{subfigure}[b]{0.3\textwidth}
         \centering
         \includegraphics[width=\textwidth]{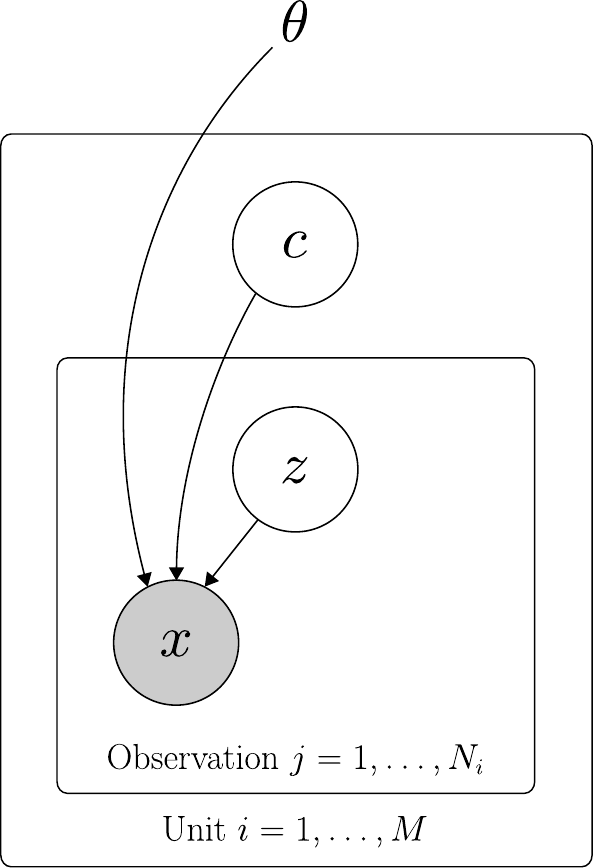}
         \caption{ML-VAE}
         \label{fig:ml-vae}
     \end{subfigure}
     \hfill
     \begin{subfigure}[b]{0.3\textwidth}
         \centering
         \includegraphics[width=\textwidth]{figs/graphical-models/gen_model_v4.pdf}
         \caption{SSMTL-VAE}
         \label{fig:SSMTL}
     \end{subfigure}
        \caption{Related generative models}
        \label{fig:gen-models}
\end{figure}

\paragraph{VAE for unsupervised learning.}
The basic VAE assumes the following generative model:
\begin{equation*}
    p_{\theta}(x, z) = p_{\theta}(x \condbar z) p(z),
\end{equation*}
where $x$ is the data and $z$ is a latent variable \cite{Kingma2019}. A simple normal prior $p(z) = \normaldist(0, I)$ for the latent variables is combined with an expressive conditional distribution, $p_{\theta}(x \condbar z)$. Expressiveness is obtained by parameterizing the conditional distribution by a neural network. The generative model, $p_{\theta}(x, z)$, is thus a deep latent variable model (DLVM). The VAE is used for unsupervised learning since the marginal distribution of $x$ is approximated by $p_{\theta}(x) = \int p_{\theta}(x, z) dz$.

\paragraph{VAE for semi-supervised learning.}
The VAE was first applied to semi-supervised classification by \cite{Kingma2014b}. The generative model of the VAE was extended to include an occasionally observed label $y$. The resulting model, originally named M2, assumed the following process $z \rightarrow x \leftarrow y$.

Several works have later proposed different extensions of the VAE to enable semi-supervised learning; cf. \cite{Yang2021}. Figure \ref{fig:ssl-vae} shows one extension where the data is modeled by a common latent variable $z$. The model, which assumes the generative process $x \leftarrow z \rightarrow y$, was used in the soft sensing method of \cite{Xie2020}.  




\paragraph{Multi-level VAE for grouped data.}
The VAE assumes i.i.d. data and is not suitable for grouped data. To handle non-i.i.d. grouped data, the multi-level VAE (ML-VAE) adds a second level to the VAE \cite{Bouchacourt2018}. At the second level, a latent group/context variable $c$ is introduced, as shown in Figure \ref{fig:ml-vae}. The ML-VAE then models $p_{\theta}(x, z \condbar c) = p_{\theta}(x \condbar z, c) p(z)$, where the conditioning on $c$ allows the i.i.d. assumption on within-group data.

The ML-VAE can be viewed as a conditional VAE by considering that the model $p_{\theta}(x, z \condbar c)$ factorizes as above under the independence assumption ${z \independence c}$ \cite{Sohn2015}. Several variations of the multi-level VAE exist in the literature; one example is the model in \cite{Edwards2016}.

\paragraph{Semi-supervised learning on grouped data.}
As illustrated in Figure \ref{fig:gen-models}, the SSMTL model is obtained by combining the structures of the SSL-VAE and ML-VAE. The observation-level structure of the SSMTL model is similar to the SSL-VAE. And like the ML-VAE, an additional level with latent variables $c$ is introduced to model variations among groups/units. The resulting structure of the SSMTL model enables semi-supervised learning on grouped data.

\paragraph{Multi-task learning.}
The learning problem solved by the SSMTL model can also be framed as a multi-task learning problem. In this frame, the tasks are to model the data for each of the $M$ groups/units. The hierarchical structure of the SSMTL model allows commonalities among tasks to be captured by the $\theta$ parameters, while task specificities are modeled by the context variables, $c$. In the MTL framework, the tasks are simultaneously solved, as in the inference method of the SSMTL model.

\begin{figure}[bt]
\centering
\includegraphics[width=0.4\linewidth]{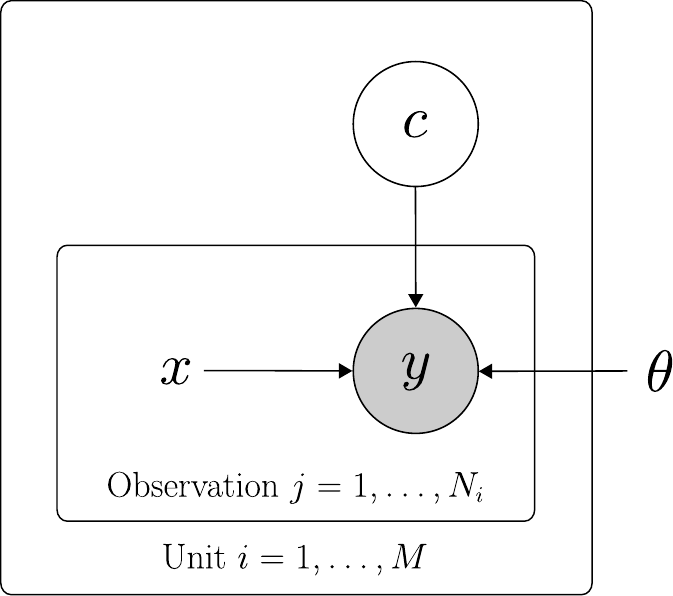}
\caption{Discriminative MTL model}
\label{fig:mtl-model}
\end{figure}

In the following numerical study, we will compare our model to a simpler model for supervised MTL. The MTL model, shown in Figure \ref{fig:mtl-model}, is given as $p_{\theta}(y \condbar x, c)p(c)$. This is a discriminative model since $p(x)$ is not modeled.

\section{Case study: Virtual Flow Metering}
\label{sec:empirical-study}
In this section, the proposed SSMTL method is evaluated on the soft sensing problem of \textit{Virtual Flow Metering} (VFM), which is introduced in Section \ref{sec:vfm}. In Section \ref{sec:datasets}, the datasets used in the case study are described. Experimental results are presented in Section \ref{sec:experiments}.

\subsection{Virtual Flow Metering}
\label{sec:vfm}
\begin{figure}[bt]
    \centering
    \includegraphics[width=\textwidth]{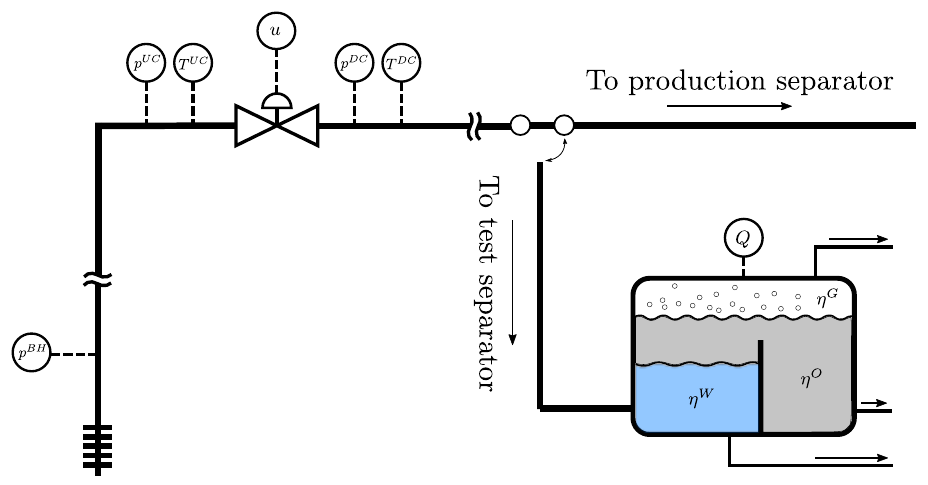}
    \caption{Petroleum production system subject to well testing. The multi-phase flow of produced liquid is either routed directly to a processing facility without being measured, or to a test separator which splits the flow into three single-phase flows which are easier to measure. The observations of pressure, temperature and choke opening are continuously available, while measurements of flow rate and phase fractions are scarce due to the cost of well testing.}
    \label{fig:well-sensors}
\end{figure}

In the petroleum industry, many critical decisions rely on measurements of flow. However, the flow from petroleum wells consists of multiple phases (oil, gas and water), which is difficult to measure precisely without first separating the phases. Since flow from multiple wells are typically combined before reaching the processing facility where this is done, information about flow from individual wells is often unavailable. The traditional method for performing multi-phase flow rate measurements in petroleum wells is illustrated in Figure \ref{fig:well-sensors}, and consists of routing flow to a test separator where gravity separates the phases into three single-phase flows, each of which can be reliably measured \cite{corneliussen_2005_handbook}. This method, which is called \textit{well testing}, requires a re-routing intervention which tends to decrease production. Well tests are consequently conducted quite rarely, typically every 7-60 days. Noninvasive physical measurement devices known as \textit{Multi-Phase Flow Meters} (MPFM) do exist, and can be installed on individual wells. However, these are typically costly, while also being less precise, requiring regular maintenance, and suffering from sensor drift over time \cite{falcone2001multiphase}.

In summary, well flow measurements are often scarce. Measurements like pressure and temperature, however, are generally abundant. This motivates the application of a soft sensor. In the context of well flow estimation, soft sensors are called \textit{Virtual Flow Meters} (VFM).
The VFM literature has mainly been concerned with discriminative models, which have been shown to work well in many settings \cite{Bikmukhametov2020}.
However, most of the available data from wells are typically unsupervised due to the low frequency of well tests. As a consequence, current data-driven VFM approaches leave large amounts of the available data unused.

Thus, the VFM problem aligns well with the motivation for semi-supervised learning presented in Section \ref{sec:introduction}. In the remainder of this section we investigate whether our proposed SSMTL methodology can leverage these unlabeled data, improving data-driven VFM performance. We do this both on a simple, simulated flow estimation problem, and on a large dataset containing production data from real petroleum wells.

In settings with real multi-unit well data, models based on MTL have repeatedly been shown to outperform competing methods, both purely data-driven and physics-informed VFMs \cite{Sandnes2021, Hotvedt2022, grimstad2023multi}. Based on this, we choose to benchmark our proposed method against  a discriminative MTL method like the one which is illustrated in Figure \ref{fig:mtl-model}. To investigate the significance of the multi-unit nature of the data, we also consider a simple strategy for single-task learning (STL), where an independent model is trained for each unit.

\subsection{Datasets}
\label{sec:datasets}

\paragraph{Synthetic data}
Any semi-supervised learning algorithm can fail to yield performance improvements over supervised methods if the learning problem at hand does not lend itself to semi-supervised learning \cite{Seeger2006, Scholkopf2012}. Thus, it is useful to evaluate the proposed SSMTL method on a learning problem where one can be confident that unlabeled data can be of use. Motivated by this, a simulated system inspired by the VFM problem was considered. Just as the petroleum production system which is illustrated in Figure \ref{fig:well-sensors}, this system consists of a flow from a pressurized reservoir, and the soft sensing objective is to predict the flow rate $Q$ given other available measurements. In this case the available measurements are a single pressure measurement $p$ and a percent-wise choke valve opening $u$, meaning the regressor and target are defined as follows:
\begin{align}
    x_{ij} & = [u_{ij}, p_{ij}] \\
    y_{ij} & = Q_{ij}
\end{align}
In contrast to the real petroleum production system, the simulated flow only consists of a single fluid, and it is modelled by simple equations for flow with friction through a vertical pipe ending in a choke valve. By generating data from simulated units with different physical parameters, the model can be learned across units by leveraging MTL methodology. This idealized setting, where data for each unit are i.i.d. and have no observation noise, is suited to inform us whether the proposed SSMTL can work, given that the problem admits semi-supervised learning.

\begin{figure}[bt]
    \centering
    \includegraphics[width=\textwidth]{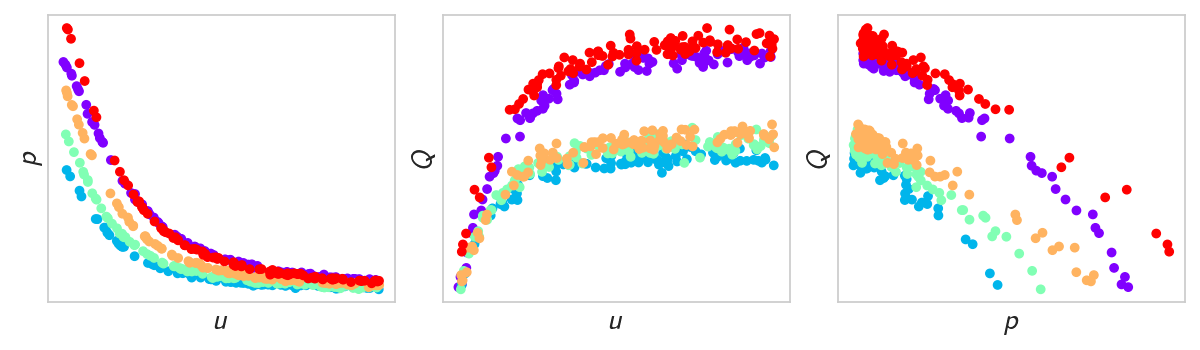}
    \caption{Synthetic data from five different units. Data points from the same unit have the same color.}
    \label{fig:synthetic-data}
\end{figure}

An example of synthetic data is shown in Figure \ref{fig:synthetic-data}. From this figure one can see that the data from the different units are qualitatively similar, but quantitatively different. This motivates the use of a hierarchical model for grouped data or a multi-task learning method.

\paragraph{Real data}
To evaluate the real-life utility of the proposed modelling approach, a dataset consisting of historical production data from real petroleum wells was considered. For this dataset, the regressor and target are defined as
\begin{align}
    x_{ij} & = [u_{ij}, p^{\textup{UC}}_{ij}, p^{\textup{DC}}_{ij}, T^{\textup{UC}}_{ij}, \eta^{\textup{OIL}}_{ij}, \eta^{\textup{GAS}}_{ij}, Q^{\textup{GL}}_{ij}] \\
    y_{ij} & = Q^{\textup{TOT}}_{ij}
\end{align}
were $u$ is the choke valve opening, $p^{\textup{UC}}$ and $p^{\textup{DC}}$ are pressures upstream and downstream the choke valve, $T^{\textup{UC}}$ is temperature upstream the choke valve, $\eta^{\textup{OIL}}$ and $\eta^{\textup{GAS}}$ are volumetric fractions of oil and gas, $Q^{\textup{GL}}$ is volumetric gas lift rate, and $Q^{\textup{TOT}}$ is total volumetric flow rate. All of these measurements are illustrated in Figure \ref{fig:well-sensors}. For a more in-depth discussion of this system and the VFM problem in general, see \cite{Bikmukhametov2020}.

In the dataset, each data point is calculated as an average over a period of steady-state operation, detected using the technology described in \cite{Grimstad2016}. Thus, we only model steady-state flow, as is common in the VFM literature \cite{Bikmukhametov2020}. 

As illustrated in Figure \ref{fig:well-sensors}, $\eta^{\textup{OIL}}$ and $\eta^{\textup{GAS}}$ arise from from the three single-phase flow measurements which together make up the multi-phase flow measurement. Thus, these will be subject to the same data limitations as the label $y = Q^{\textup{TOT}}$. To account for this, we assume that fractions change slowly over time, and approximate them using their value from the previous available flow measurement.

\begin{figure}[bt]
    \centering
    \includegraphics[width=\textwidth]{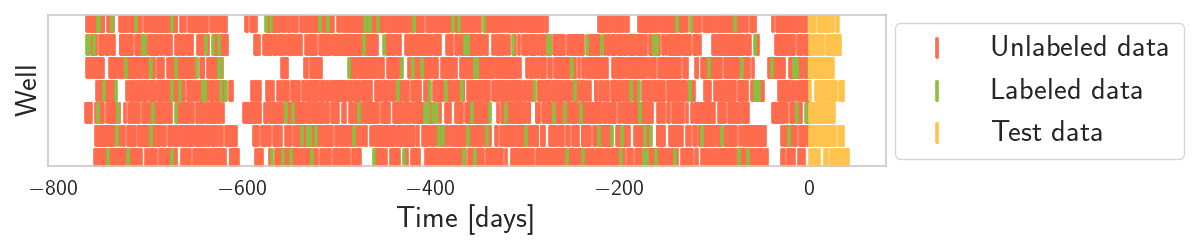}
    \caption{Data used for the multi-unit semi-supervised learning experiment. This figure shows data from seven of the 64 wells, for the ratio $N^u / N^l = 20$ of unlabeled to labeled data. A total of 1280 labeled, 25600 unlabeled and 1280 test data points are available, distributed equally among the units. A rectangular marker indicates the availability of a data point. The x-axis is shifted for each unit such that $t=0$ coincides with the start of the test set.}
    \label{fig:real-well-data}
\end{figure}

\begin{figure}[bt]
    \centering
    \includegraphics[width=\textwidth]{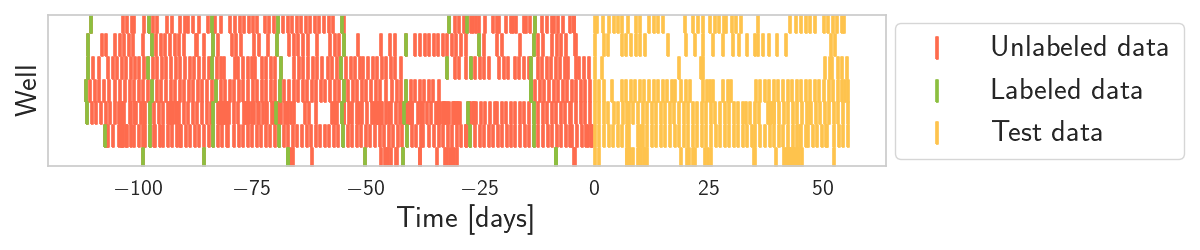}
    \caption{Data used for the finetuning experiment. Across all units, a total of 56 labeled and 589 unlabeled and 283 data points are available. The labeled data points are equally distributed among the units, while the unlabeled and test data are not. The ratio $N^u / N^l$ of unlabeled to labeled data varies between units, and across time intervals; on average, it is about 10. A rectangular marker indicates the availability of a data point. The x-axis is shifted for each unit such that $t=0$ coincides with the start of the test set.}
    \label{fig:real-well-data-few-shot}
\end{figure}

Accurate evaluation of model generalization requires access to sufficient amounts of labeled test data. This is hard to achieve in the label-scarce setting described in Section \ref{sec:vfm}. To gain trustworthy estimates of model generalization error, we therefore make use of data from wells where both well tests and MPFM measurements are available. For these wells, labeled data are available at the same rate as unlabeled data, and we can generate sizeable test sets. To reproduce the data scarcity which motivates this work (and which is very real on poorlier instrumented wells), we then remove the labels from the majority of the training data. To mimic the setting in which a VFM would be used in practice, the split into training and test set is done chronologically. The result is a dataset where test data and unlabeled training data are abundant, while labeled training data are scarce. Figure \ref{fig:real-well-data} and Figure \ref{fig:real-well-data-few-shot} illustrate the availability of the different types of data for the datasets which are used in Section \ref{sec:experiments}.

\subsection{Experiments}
\label{sec:experiments}
For both datasets, we investigate model performance in two data availability regimes: 1) a \textit{multi-unit learning} setting where partly labeled data from many units are available, and 2) a \textit{single-unit finetuning} setting where the same multi-unit data are available, but where the objective is to model a previously unseen unit with little data. In both of these settings, our goal is to answer the following question: Can our proposed SSMTL model use unlabeled data to improve soft sensor accuracy?

\paragraph{Multi-unit learning}
The first case study addresses the abilities of the models to learn from partly labeled multi-unit data. Here, models based on single-task learning (STL), multi-task learning (MTL) and semi-supervised multi-task learning (SSMTL) are considered.

For this experiment, synthetic data were generated for 50 different units, each unit having different physical constants like pipe height, pipe diameter and friction coefficient. For each of the units, a training dataset with 5 labeled and 100 unlabeled data points was generated, in addition to a test set with 100 data points for each unit. The real dataset consisted of historical data from 64 wells, where 20 labeled and 400 unlabeled and 20 test data points were available for each well. For both datasets the effect of increasing the amount of unlabeled data was investigated by considering three different unlabeled datasets, where the ratio $N^u / N^l$ of unlabeled to labeled data was 1, 5 and 20, respectively. The real dataset for $N^u / N^l = 20$ is illustrated for seven of the 64 real wells in Figure \ref{fig:real-well-data}.

The SSMTL method was trained using the SGVB estimator of Algorithm \ref{alg:sgvb}, while the MTL and STL models were trained using stochastic gradient descent on the log-likelihood objective. Early stopping was used to decide which iteration of the training to use for model evaluation. To reduce variance, results for the MTL and SSMTL models were generated by repeating the experiment 20 times with different random initialization of the neural networks, and averaging results for each unit over experiment repetitions. For further implementation details, see Appendix \ref{app:implementation-details}.

Table \ref{tab:results-synthetic-case} and \ref{tab:results-real-case} show the mean absolute percentage error (MAPE) on the test set for the synthetic and real dataset, respectively.
For both synthetic and real data, a significant decrease in error is achieved when going from STL to MTL.
The performance improvement when going from MTL to SSMTL is comparatively lower, but still evident in both cases. The improvement is consistent across the synthetic and real data, and across different amounts of unlabeled data.
For the synthetic data, model performance also increases as the amount of unlabeled data increases. For the real data, this does not seem to be the case. The performance improvement when going from zero to some unlabeled data, is however evident on both cases.

\begin{table}[bt]
\centering
\caption{Prediction error on synthetic test data. The average, the 10th, the 50th and the 90th percentile of the MAPE is reported. The lowest value of each error statistic is set in boldface.}
\label{tab:results-synthetic-case}
\begin{tabularx}{\textwidth}{
    l
    >{\centering\arraybackslash}X 
    >{\centering\arraybackslash}X 
    >{\centering\arraybackslash}X
    >{\centering\arraybackslash}X
    >{\raggedleft\arraybackslash}X }
    \toprule
    Model & Unlabeled ratio & Mean & $P_{10}$ & $P_{50}$ &  $P_{90}$ \\
    \midrule
    STL & - & 22.302 & 15.638 & 22.356 & 27.894 \\
    MTL & - & 3.671 & 2.676 & 3.374 & 5.076 \\
    SSMTL & 1 & 3.534 & 2.720 & 3.363 & 4.380 \\
    SSMTL & 5 & 3.375 & 2.602 & 3.208 & \textbf{4.177} \\
    SSMTL & 20 & \textbf{3.310} & \textbf{2.434} & \textbf{3.178} & 4.311 \\
    \bottomrule
\end{tabularx}
\end{table}
\begin{table}[bt]
\centering
\caption{Prediction error on real test data. The average, the 10th, the 50th and the 90th percentile of the MAPE is reported. The lowest value of each error statistic is set in boldface.}
\label{tab:results-real-case}
\begin{tabularx}{\textwidth}{
    l
    >{\centering\arraybackslash}X 
    >{\centering\arraybackslash}X 
    >{\centering\arraybackslash}X
    >{\centering\arraybackslash}X
    >{\raggedleft\arraybackslash}X }
    \toprule
    Model & Unlabeled ratio & Mean & $P_{10}$ & $P_{50}$ &  $P_{90}$ \\
    \midrule
    STL & - & 14.726 & 4.416 & 11.620 & 27.059 \\
    MTL & - & 11.737 & 3.114 & 7.219 & 25.729 \\
    SSMTL & 1 & 10.815 & 3.068 & \textbf{6.729} & 24.917 \\
    SSMTL & 5 & 10.774 & \textbf{2.887} & 6.887 & 24.908 \\
    SSMTL & 20 & \textbf{10.679} & 3.433 & 7.182 & \textbf{24.741} \\
    \bottomrule
\end{tabularx}
\end{table}

\paragraph{Single-unit calibration}
The second case study considers a setting where a previously unseen unit is to be modelled. For this unit, data points are appearing incrementally, and can be used to update the model which was learned in the previous case study in an online manner. If the base training set consists of data from sufficiently many units, one can motivate a simplified learning procedure wherein the neural network parameters are frozen, and only the context parameters are learned. This method was used here. For further motivation of this methodology, see \cite{grimstad2023multi}.

For the synthetic dataset, data from 20 new units were generated. For each unit, $N^l$ labeled and $N^u = 10 N^l$ unlabeled observations were available for $N^l = 1, 2, \dots, 10$ labeled observations, in addition to 100 test data points for each unit. For the real dataset, data from seven units which were not part of the multi-unit learning experiment were used. To mimic a setting where new data are incrementally observed, a test set with a duration of eight weeks was fixed for each unit, while labeled data points were chosen to appear every other week, incrementally and backwards in time, for $N^l = 1, 2, \dots 8$. The number of unlabeled data points was only limited by the original dataset. For the largest number of weeks considered (16 weeks), a total of 56 labeled, 589 unlabeled and 283 test data points were available, meaning the average unlabeled to labeled ratio $N^u / N^l$ was approximately 10. Data availability in this case is illustrated in Figure \ref{fig:real-well-data-few-shot}.

For each new labeled datapoint that appeared (as well as its associated $\approx 10$ unlabeled data points), a pre-trained SSMTL model was finetuned to the available data. This was done both in a supervised (only labeled data), an unsupervised (only unlabeled data) and a semi-supervised (both labeled and unlabeled data) fashion. In each case, learning was done by taking a fully trained model from the multi-unit learning experiment, freezing its neural network parameters, and learning the parameters of a freshly initialized context module from the data available for the given unit. For further details, see Appendix A.

Figure \ref{fig:few-shot-results} shows the resulting model performance. In both the synthetic and real data case, the model is able to learn from unsupervised data; test set error decreases as purely unlabeled data become available (red line), and when labeled data are available, the SSMTL model which has access to unlabeled data (blue line) consistently outperforms the one which does not (green line).

\begin{figure}[bt]
    \centering
    \begin{subfigure}[b]{0.49\textwidth}
        \centering
        \includegraphics[width=\textwidth]{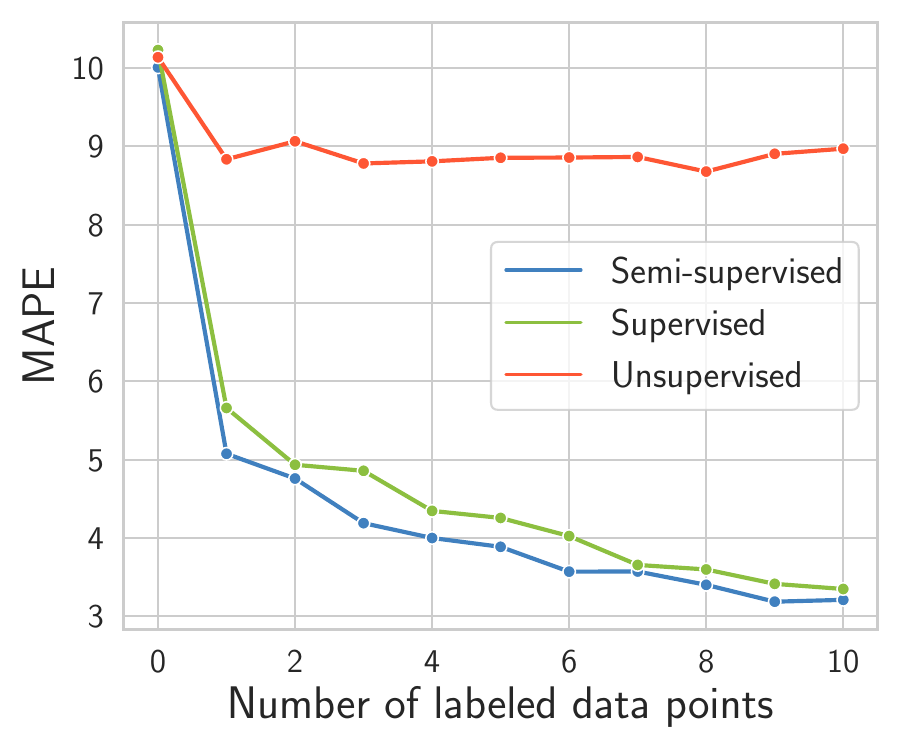}
        \caption{Synthetic data}
        \label{fig:few-shot-results-synthetic-data}
    \end{subfigure}
    \begin{subfigure}[b]{0.49\textwidth}
        \centering
        \includegraphics[width=\textwidth]{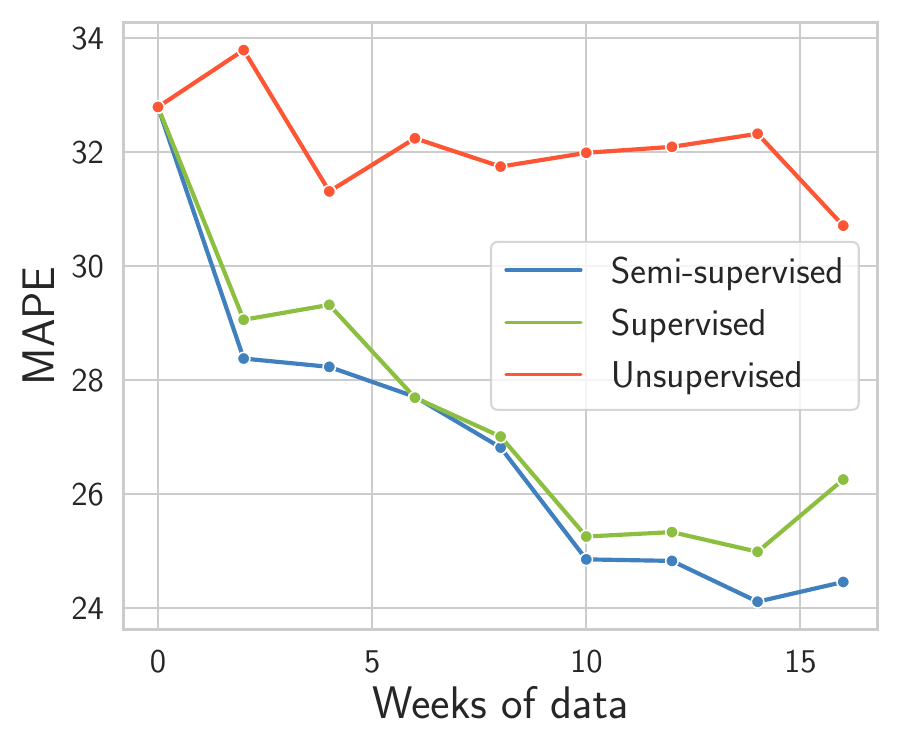}
        \caption{Real data}
        \label{fig:few-shot-results-real-data}
    \end{subfigure}
    \caption{Finetuned model performance on real and synthetic data. For the synthetic data case, the ratio of unlabeled to labeled data is $N^u / N^l = 10$. For the real data case, the ratio varies between wells, but is also, on average, close to 10.}
    \label{fig:few-shot-results}
\end{figure}

\section{Discussion}
\label{sec:discussion}
The results presented in Section \ref{sec:empirical-study} suggest that the proposed SSMTL model is able to learn from unlabeled data, improving performance in the soft sensing problem of data-driven VFM. This conclusion holds across datasets and case studies: unlabeled data improve performance for classical multi-unit modelling and single-unit finetuning, both for synthetic and real data.

In terms of pure numbers, the performance improvement gained from semi-supervised learning in the case study considered here is arguably quite modest. However, we emphasize that the question of interest to us is whether unlabeled data can be helpful at all; the degree of performance improvement which is attainable is limited by the learning problem and the data at hand \cite{Seeger2006, Scholkopf2012}.

Considering the fact that VAEs have proven capable of modelling high-dimensional data \cite{Kingma2019}, we expect that a further performance improvement from semi-supervised learning can be gained in settings with more complex, high-dimensional regressors, for instance settings with additional sensors, sensors that provide higher-dimensional features, or for modelling sequences of observations. All of these model extensions provide interesting avenues for further research.

In light of this, it is remarkable that the SSMTL method is able to achieve a measurable performance increase at all in the setting considered here, where the regressor $x$ has dimension 7, each feature being selected to be as informative as possible regarding $y$. We also emphasize the fact that in the single-unit calibration case study, a performance improvement was seen even in purely unsupervised settings, where \textit{no labels} were available.

\section{Concluding remarks}
\label{sec:conclusion}
We have proposed a deep probabilistic model which performs semi-supervised learning from multi-unit data, and demonstrated on real data from a significant industrial problem that the model can use unlabeled data to improve soft sensor performance. The proposed model provides a general approach to multi-unit soft sensing problems, and by combining the advantages of multi-task and semi-supervised learning, the method is able to learn from very few data points. This makes it is especially powerful in information-poor environments, where data efficiency is essential. 

\section*{Acknowledgements}
This work was supported by Solution Seeker Inc. 

\appendix

\section{Implementation details}
\label{app:implementation-details}

\paragraph{STL models.} Single-task learning was performed by training one model per unit in the datasets. The models were trained to minimize the mean squared error on labeled data only. For these tasks, we used support vector regression with a radial basis function kernel \cite{Smola2004}. To avoid overfitting, the $L^2$-regularization factor was found by performing a hyperparameter optimization on randomly selected validation data.

\paragraph{MTL models.} The MTL model was implemented using a feed-forward neural network. The likelihood of an observation $(x_{ij}, y_{ij})$ from unit $i$ was modeled as $y_{ij} \sim \normaldist(f_{\theta}(x_{ij}, c_{i}), \sigma^{2})$, where $f_{\theta}$ is a neural network with parameters $\theta$. The context variables were modeled by a standard normal prior and a variational distribution $c_i \sim \normaldist(m_i, \text{diag}(s_{i}^{2}))$ for $i=1,\ldots,M$. The parameters $\theta$, $\sigma$, $\{m_i\}_i$, and $\{s_i\}_i$ were learned from data. Note that the $\theta$ and $\sigma$ parameters were shared by all units. In the synthetic data case, $f_{\theta}$ had two hidden layers, each with 200 nodes. In the real data case, $f_{\theta}$ had two hidden layers, each with 400 nodes. The dimension of the context variable was $K=4$ in both cases.

To regularize the MTL model, $L_2$ regularization was also implemented. However, initial experiments indicated that the highest performance of the MTL model was achieved without explicit neural network parameter regularization (we hypothesize that this is due to early stopping). Consequently, $L_2$ regularization was not used in the case studies presented here.

\paragraph{SSMTL models.} The proposed model for semi-supervised multi-task learning (SSMTL) was implemented using three feed-forward neural networks: one for the likelihood, $p_{\theta}(x, y \condbar z, c)$, and two for the inference models, $q_{\phi_z}(z \condbar x, y, c)$ and $q_{\phi_y}(y \condbar x, c)$. For the synthetic data case, all four networks had two hidden layers, each with 200 nodes. For the real data case, all four networks had two hidden layers, each with 400 nodes. Variable dimensions used in the two cases are shown in Table \ref{tab:variable-dimensions}. Notice that we used an overcomplete VAE with $D > D_x + D_y$, where regularization is applied via the prior on the latent variables.

On the synthetic data case, we set $\alpha = 1$. On the real data case, we found that the performance was improved by setting $\alpha = 10$, putting more weight on log-likelihood of the inference model on labeled data.

\begin{table}[bt]
\centering
\caption{Dimensions of variables in the SSMTL model for the synthetic and real data case.}
\label{tab:variable-dimensions}
\begin{tabularx}{\textwidth}{
    l
    >{\centering\arraybackslash}X 
    >{\centering\arraybackslash}X
    >{\centering\arraybackslash}X
    >{\raggedleft\arraybackslash}X }
    \toprule
    Case & $K$ & $D$ & $D_x$ & $D_y$ \\
    \midrule
    Synthetic data & 4 & 5 & 2 & 1 \\
    Real data & 4 & 10 & 7 & 1 \\
    \bottomrule
\end{tabularx}
\end{table}

\paragraph{Model architecture and optimization.} All neural networks were implemented in PyTorch \cite{Pytorch2019}. We used ReLU activation functions and initialized the parameters using He initialization \cite{He2015}.

For the multi-unit learning case study, we used the Adam optimizer \cite{Kingma2015} to train the SSMTL and MTL models. We used a five-sample SGVB estimate for the SSMTL model due to the additional variance introduced by the $z$ variable. The learning rate was manually tuned to $5 \times 10^{-4}$. The other hyperparameters of Adam were kept at their standard values. We used early stopping on randomly selected validation data to prevent overfitting.

For the single-unit finetuning experiment, plain stochastic gradient descent (SGD) was used. Based on manual tuning, 100 epochs of SGD with a learning rate of $10^{-4}$ were performed for each unit, for all models. For every new data point that became available, training was done from scratch. For the synthetic data, context parameter means were initialized at their prior mean, i.e. at zero. For real data, the context parameter mean was set to the average of context parameter means on the same field. For both datasets, the variance of $q_{\varphi}(c_i)$ was fixed at $0.1$.

All models were trained on a desktop computer with a single GPU.

\section{Generation of synthetic data}
\label{sec:synthetic-dataset-generation}
We base the generative process of our synthetic data on the on the following equations which, respectively, model single-phase fluid flow through a choke valve and pressure drop in a vertical pipe with friction:
\begin{align}
    Q & = C u \sqrt{\frac{2 (p - p_o)}{\rho}} \\
    p & = p_r - \rho g h - f_D \frac{8 \rho h}{\pi^2 D^5}Q^2
\end{align}
Here, Q is the single-phase flow rate, $u \in [0, 1]$ is the choke valve position, $p$ is the pressure upstream the choke valve, $C$ is the choke CV coefficient, $\rho$ is the fluid density, $p_o$ is the outlet pressure, $p_r$ is the bottomhole pressure, $g$ is the gravitational acceleration, $h$ is the height of the pipe, $f_D$ is the Darcy friction coefficient, an $D$ is the diameter of the pipe. $Q$, $u$ and $p$ are measured variables, while all other values entering the equations are parameters.

The parameter space can be simplified by combining all parameters which enter together. We define constants
\begin{equation*}
    c_1 = \frac{2 C^2}{\rho} p_o,\quad c_2 = \frac{2 C^2}{\rho}, \quad c_3 = p_r - \rho g h ,\quad c_4 = f_D \frac{8 \rho h}{\pi^2 D^5}.
\end{equation*}
The equations can then be written as
\begin{align}
    Q & = \sqrt{-c_1 u^2 + c_2 u^2 p} \\
    p & = c_3 - c_4 Q^2
\end{align}
Gaussian noise is added to measurements of $p$ and $Q$. Pseudocode for the data generating process is given below.

\begin{algorithm}[bt]
\caption{Synthetic data generation}
\label{alg:synth-data-generation}
\begin{algorithmic}[1]
\Require Parameter distribution $p(c_1, c_2, c_3, c_4)$. 
\State Draw parameters $c_{1}$, $c_{2}$, $c_{3}$, $c_{4}$ 
\For{$i=1,\ldots,M$}
\For{$j=1, \ldots,N_i$}
\State $u_{ij} \sim \uniformdist(0, 100)$
\State $\mu_p \gets \frac{c_3 + c_1 c_4 u_{ij}^2}{1 + c_2 c_4 u_{ij}^{2}}$
\State $p_{ij} \sim \normaldist(\mu_p, \sigma_{p}^{2})$
\State $\mu_Q \gets \sqrt{- c_1 u_{ij}^2 + c_2 u_{ij}^{2} \mu_p}$
\State $Q_{ij} \sim \normaldist(\mu_Q, \sigma_{Q}^{2})$
\EndFor
\EndFor
\State \Return $u, p, Q$
\end{algorithmic}
\end{algorithm}

\FloatBarrier
\printbibliography

@article{Blei2003,
author = {Blei, David M. and Ng, Andrew Y. and Jordan, Michael I.},
isbn = {9780124115439},
journal = {Journal of Machine Learning Research},
pages = {993--1022},
title = {{Latent Dirichlet Allocation}},
volume = {3},
year = {2003}
}

@article{Yang2021,
archivePrefix = {arXiv},
arxivId = {2103.00550},
author = {Yang, Xiangli and Song, Zixing and King, Irwin and Xu, Zenglin},
eprint = {2103.00550},
pages = {1--26},
title = {{A Survey on Deep Semi-supervised Learning}},
url = {http://arxiv.org/abs/2103.00550},
year = {2021}
}

@article{Esche2022,
author = {Esche, Erik and Talis, Torben and Weigert, Joris and {Brand Rihm}, Gerardo and You, Byungjun and Hoffmann, Christian and Repke, Jens Uwe},
doi = {10.1016/j.ces.2022.117459},
eprint = {2107.13822},
issn = {00092509},
journal = {Chemical Engineering Science},
pages = {117459},
publisher = {Elsevier Ltd},
title = {{Semi-supervised learning for data-driven soft-sensing of biological and chemical processes}},
volume = {251},
year = {2022}
}

@article{Kadlec2009,
author = {Kadlec, Petr and Gabrys, Bogdan and Strandt, Sibylle},
doi = {10.1016/j.compchemeng.2008.12.012},
issn = {00981354},
journal = {Computers and Chemical Engineering},
number = {4},
pages = {795--814},
title = {{Data-driven Soft Sensors in the process industry}},
volume = {33},
year = {2009}
}

@inproceedings{Liu2007,
author = {Liu, Qiuhua and Liao, Xuejun and Carin, Lawrence},
booktitle = {Advances in Neural Information Processing Systems 20},
title = {{Semi-Supervised Multitask Learning}},
year = {2007}
}

@article{Xie2020,
author = {Xie, Ruimin and Jan, Nabil Magbool and Hao, Kuangrong and Chen, Lei and Huang, Biao},
doi = {10.1109/TII.2019.2951622},
issn = {1551-3203},
journal = {IEEE Transactions on Industrial Informatics},
number = {4},
pages = {2820--2828},
title = {{Supervised Variational Autoencoders for Soft Sensor Modeling With Missing Data}},
volume = {16},
year = {2020}
}

@article{Sun2021,
author = {Sun, Qingqiang and Ge, Zhiqiang},
doi = {10.1109/TII.2021.3053128},
issn = {1551-3203},
journal = {IEEE Transactions on Industrial Informatics},
number = {9},
pages = {5853--5866},
publisher = {IEEE},
title = {{A Survey on Deep Learning for Data-Driven Soft Sensors}},
url = {https://ieeexplore.ieee.org/document/9329169/},
volume = {17},
year = {2021}
}

@article{Jiang2021,
author = {Jiang, Yuchen and Yin, Shen and Dong, Jingwei and Kaynak, Okyay},
doi = {10.1109/JSEN.2020.3033153},
issn = {1530-437X},
journal = {IEEE Sensors Journal},
number = {11},
pages = {12868--12881},
title = {{A Review on Soft Sensors for Monitoring, Control, and Optimization of Industrial Processes}},
volume = {21},
year = {2021}
}

@article{Kingma2019,
archivePrefix = {arXiv},
arxivId = {1906.02691},
author = {Kingma, Diederik P. and Welling, Max},
doi = {10.1561/2200000056},
eprint = {1906.02691},
issn = {19358245},
journal = {Foundations and Trends in Machine Learning},
number = {4},
pages = {307--392},
title = {{An introduction to variational autoencoders}},
volume = {12},
year = {2019}
}

@inproceedings{Edwards2016,
author = {Edwards, Harrison and Storkey, Amos},
booktitle = {5th International Conference on Learning Representations},
eprint = {1606.02185},
pages = {1--13},
title = {{Towards a Neural Statistician}},
year = {2017}
}

@inproceedings{Zhang2009,
author = {Zhang, Yu and Yeung, Dit-yan},
booktitle = {Joint European Conference on Machine Learning and Knowledge Discovery in Databases},
doi = {10.1007/978-3-642-04174-7_40},
pages = {617--631},
title = {{Semi-Supervised Multi-Task Regression}},
year = {2009}
}

@incollection{Seeger2006,
author = {Seeger, Matthias},
booktitle = {Semi-Supervised Learning},
chapter = {2},
editor = {Chapelle, O. and Schoelkopf, B. and Zien, A.},
pages = {15--32},
publisher = {MIT Press},
title = {{A Taxonomy for Semi-Supervised Learning Methods}},
year = {2006}
}

@inproceedings{Zhuang2015,
author = {Zhuang, Fuzhen and Luo, Dan and Jin, Xin and Xiong, Hui and Luo, Ping and He, Qing},
booktitle = {IEEE International Conference on Data Mining},
isbn = {978-1-4673-9504-5},
issn = {15504786},
pages = {1141--1146},
publisher = {IEEE},
title = {{Representation Learning via Semi-Supervised Autoencoder for Multi-task Learning}},
year = {2015}
}

@article{Shen2020,
author = {Shen, Bingbing and Yao, Le and Ge, Zhiqiang},
doi = {10.1016/j.conengprac.2019.104198},
issn = {09670661},
journal = {Control Engineering Practice},
number = {March 2019},
pages = {104198},
publisher = {Elsevier Ltd},
title = {{Nonlinear probabilistic latent variable regression models for soft sensor application: From shallow to deep structure}},
volume = {94},
year = {2020}
}

@article{VanEngelen2020,
author = {van Engelen, Jesper E. and Hoos, Holger H.},
doi = {10.1007/s10994-019-05855-6},
issn = {15730565},
journal = {Machine Learning},
number = {2},
pages = {373--440},
publisher = {Springer US},
title = {{A survey on semi-supervised learning}},
volume = {109},
year = {2020}
}

@article{Scholkopf2012,
author = {Sch{\"{o}}lkopf, Bernhard and Janzing, Dominik and Peters, Jonas and Sgouritsa, Eleni and Zhang, Kun and Mooij, Joris},
eprint = {1206.6471},
isbn = {9781450312851},
journal = {Proceedings of the 29th International Conference on Machine Learning, ICML 2012},
pages = {1255--1262},
title = {{On causal and anticausal learning}},
volume = {2},
year = {2012}
}

@inproceedings{Kingma2014a,
archivePrefix = {arXiv},
arxivId = {1312.6114},
author = {Kingma, Diederik P and Welling, Max},
booktitle = {2nd International Conference on Learning Representations (ICLR)},
eprint = {1312.6114},
title = {{Auto-Encoding Variational Bayes}},
year = {2014}
}

@article{Kingma2014b,
author = {Kingma, Diederik P. and Rezende, Danilo J. and Mohamed, Shakir and Welling, Max},
eprint = {1406.5298},
issn = {10495258},
journal = {Advances in Neural Information Processing Systems},
number = {January},
pages = {3581--3589},
title = {{Semi-supervised learning with deep generative models}},
volume = {4},
year = {2014}
}

@article{Curreri2021,
author = {Curreri, Francesco and Patan{\`{e}}, Luca and Xibilia, Maria Gabriella},
doi = {10.3390/app11167710},
issn = {20763417},
journal = {Applied Sciences},
number = {16},
pages = {1--18},
title = {{Soft sensor transferability: A survey}},
volume = {11},
year = {2021}
}

@article{Sandnes2021,
author = {Sandnes, Anders T. and Grimstad, Bjarne and Kolbj{\o}rnsen, Odd},
doi = {10.1016/j.knosys.2021.107458},
eprint = {2103.08713},
issn = {09507051},
journal = {Knowledge-Based Systems},
pages = {107458},
publisher = {Elsevier B.V.},
title = {{Multi-task learning for virtual flow metering}},
volume = {232},
year = {2021}
}

@article{Liu2019,
author = {Liu, Yi and Yang, Chao and Liu, Kaixin and Chen, Bocheng and Yao, Yuan},
doi = {10.1016/j.chemolab.2019.103813},
issn = {18733239},
journal = {Chemometrics and Intelligent Laboratory Systems},
number = {April},
pages = {103813},
publisher = {Elsevier Ltd},
title = {{Domain adaptation transfer learning soft sensor for product quality prediction}},
volume = {192},
year = {2019}
}

@article{Yuan2020,
author = {Yuan, Xiaofeng and Gu, Yongjie and Wang, Yalin and Yang, Chunhua and Gui, Weihua},
doi = {10.1109/TNNLS.2019.2957366},
issn = {21622388},
journal = {IEEE Transactions on Neural Networks and Learning Systems},
number = {11},
pages = {4737--4746},
pmid = {31880568},
publisher = {IEEE},
title = {{A Deep Supervised Learning Framework for Data-Driven Soft Sensor Modeling of Industrial Processes}},
volume = {31},
year = {2020}
}

@article{Gordon2020,
author = {Gordon, Jonathan and Hern{\'{a}}ndez-Lobato, Jos{\'{e}} Miguel},
doi = {10.1016/j.patcog.2019.107156},
issn = {00313203},
journal = {Pattern Recognition},
pages = {107156},
publisher = {Elsevier Ltd},
title = {{Combining deep generative and discriminative models for Bayesian semi-supervised learning}},
volume = {100},
year = {2020}
}

@article{Siddharth2017,
author = {Siddharth, N. and Paige, Brooks and {Van De Meent}, Jan Willem and Desmaison, Alban and Goodman, Noah D. and Kohli, Pushmeet and Wood, Frank and Torr, Philip H.S.},
eprint = {1706.00400},
issn = {10495258},
journal = {Advances in Neural Information Processing Systems},
pages = {5926--5936},
title = {{Learning disentangled representations with semi-supervised deep generative models}},
year = {2017}
}

@article{Zhang2021,
author = {Zhang, Yu and Yang, Qiang},
doi = {10.1109/TKDE.2021.3070203},
journal = {IEEE Transactions on Knowledge and Data Engineering},
number = {c},
pages = {1--20},
title = {{A Survey on Multi-Task Learning}},
volume = {4347},
year = {2021}
}

@article{Sohn2015,
author = {Sohn, Kihyuk and Yan, Xinchen and Lee, Honglak},
issn = {10495258},
journal = {Advances in Neural Information Processing Systems},
pages = {3483--3491},
title = {{Learning structured output representation using deep conditional generative models}},
year = {2015}
}

@article{Bouchacourt2018,
author = {Bouchacourt, Diane and Tomioka, Ryota and Nowozin, Sebastian},
doi = {10.1609/aaai.v32i1.11867},
journal = {Proceedings of the AAAI Conference on Artificial Intelligence},
number = {1},
pages = {2095--2102},
title = {{Multi-Level Variational Autoencoder: Learning Disentangled Representations From Grouped Observations}},
volume = {32},
year = {2018}
}

@article{Hotvedt2022,
author = {Hotvedt, Mathilde and Grimstad, Bjarne A and Imsland, Lars S.},
doi = {10.1016/j.eswa.2022.118382},
issn = {0957-4174},
journal = {Expert Systems With Applications},
number = {February},
pages = {118382},
publisher = {Elsevier Ltd},
title = {{Passive learning to address nonstationarity in virtual flow metering applications}},
volume = {210},
year = {2022}
}

@article{Smola2004,
author = {Smola, Alex J. and Sch{\"{o}}lkopf, Bernhard},
doi = {10.1023/B:STCO.0000035301.49549.88},
journal = {Statistics and Computing},
number = {3},
pages = {199--222},
title = {{A tutorial on support vector regression}},
volume = {14},
year = {2004}
}

@article{Pytorch2019,
author = {Paszke, Adam and Gross, Sam and Massa, Francisco and Lerer, Adam and Bradbury, James and Chanan, Gregory and Killeen, Trevor and Lin, Zeming and Gimelshein, Natalia and Antiga, Luca and Desmaison, Alban and Kopf, Andreas and Yang, Edward and DeVito, Zachary and Raison, Martin and Tejani, Alykhan and Chilamkurthy, Sasank and Steiner, Benoit and Fang, Lu and Bai, Junjie and Chintala, Soumith},
journal = {Advances in Neural Information Processing Systems},
pages = {8026--8037},
title = {{PyTorch: An Imperative Style, High-Performance Deep Learning Library}},
volume = {32},
year = {2019}
}

@inproceedings{He2015,
author = {He, Kaiming and Zhang, Xiangyu and Ren, Shaoqing and Sun, Jian},
booktitle = {Proceedings of the IEEE international conference on computer vision},
doi = {10.1016/j.bbrc.2018.01.076},
pages = {1026--1034},
pmid = {7410480},
title = {{Delving Deep into Rectifiers: Surpassing Human-Level Performance on ImageNet Classification}},
year = {2015}
}

@inproceedings{Kingma2015,
author = {Kingma, Diederik P. and Ba, Jimmy Lei},
booktitle = {3rd International Conference on Learning Representations},
pages = {1--15},
title = {{Adam: A method for stochastic optimization}},
year = {2015}
}

@article{Bikmukhametov2020,
	title = {First principles and machine learning {Virtual} {Flow} {Metering}: {A} literature review},
	volume = {184},
	journal = {Journal of Petroleum Science and Engineering},
	author = {Bikmukhametov, Timur and Jäschke, Johannes},
	year = {2020},
	pages = {106487},
}

@inproceedings{Grimstad2016,
  title={A simple data-driven approach to production estimation and optimization},
  author={Grimstad, B and Gunnerud, V and Sandnes, A and Shamlou, S and Skrondal, IS and Uglane, V and Ursin-Holm, S and Foss, B},
  booktitle={SPE intelligent energy international conference and exhibition},
  pages={SPE--181104},
  year={2016},
  organization={SPE}
}

@article{sun_deep_2021,
	title = {Deep {Learning} for {Industrial} {KPI} {Prediction}: {When} {Ensemble} {Learning} {Meets} {Semi}-{Supervised} {Data}},
	volume = {17},
	issn = {1551-3203, 1941-0050},
	shorttitle = {Deep {Learning} for {Industrial} {KPI} {Prediction}},
	doi = {10.1109/TII.2020.2969709},
	language = {en},
	number = {1},
	journal = {IEEE Transactions on Industrial Informatics},
	author = {Sun, Qingqiang and Ge, Zhiqiang},
	month = jan,
	year = {2021},
	pages = {260--269},
}

@article{jin_evolutionary_2021,
	title = {Evolutionary optimization based pseudo labeling for semi-supervised soft sensor development of industrial processes},
	volume = {237},
	issn = {00092509},
	doi = {10.1016/j.ces.2021.116560},
	language = {en},
	journal = {Chemical Engineering Science},
	author = {Jin, Huaiping and Li, Zheng and Chen, Xiangguang and Qian, Bin and Yang, Biao and Yang, Jianwen},
	month = jun,
	year = {2021},
	pages = {116560},
}

@article{li_semi-supervised_2022,
	title = {Semi-supervised ensemble support vector regression based soft sensor for key quality variable estimation of nonlinear industrial processes with limited labeled data},
	volume = {179},
	issn = {02638762},
	doi = {10.1016/j.cherd.2022.01.026},
	language = {en},
	journal = {Chemical Engineering Research and Design},
	author = {Li, Zheng and Jin, Huaiping and Dong, Shoulong and Qian, Bin and Yang, Biao and Chen, Xiangguang},
	month = mar,
	year = {2022},
	pages = {510--526},
}

@article{moreira_de_lima_ensemble_2021,
	title = {Ensemble deep relevant learning framework for semi-supervised soft sensor modeling of industrial processes},
	volume = {462},
	issn = {09252312},
	doi = {10.1016/j.neucom.2021.07.086},
	language = {en},
	journal = {Neurocomputing},
	author = {Moreira De Lima, Jean Mario and Ugulino De Araujo, Fabio Meneghetti},
	month = oct,
	year = {2021},
	pages = {154--168},
}

@article{yao_semi-supervised_2023,
	title = {Semi-{Supervised} {Deep} {Dynamic} {Probabilistic} {Latent} {Variable} {Model} for {Multimode} {Process} {Soft} {Sensor} {Application}},
	volume = {19},
	issn = {1551-3203, 1941-0050},
	url = {https://ieeexplore.ieee.org/document/9797056/},
	doi = {10.1109/TII.2022.3183211},
	language = {en},
	number = {4},
	journal = {IEEE Transactions on Industrial Informatics},
	author = {Yao, Le and Shen, Bingbing and Cui, Linlin and Zheng, Junhua and Ge, Zhiqiang},
	month = apr,
	year = {2023},
	pages = {6056--6068},
}

@article{qiao_multitask_2023,
	title = {A {Multitask} {Learning} {Model} for the {Prediction} of {NOx} {Emissions} in {Municipal} {Solid} {Waste} {Incineration} {Processes}},
	volume = {72},
	issn = {0018-9456, 1557-9662},
	doi = {10.1109/TIM.2022.3225056},
	language = {en},
	journal = {IEEE Transactions on Instrumentation and Measurement},
	author = {Qiao, Junfei and Zhou, Jianglong and Meng, Xi},
	year = {2023},
	pages = {1--14},
}

@article{huang_modeling_2023,
	title = {Modeling {Task} {Relationships} in {Multivariate} {Soft} {Sensor} {With} {Balanced} {Mixture}-of-{Experts}},
	volume = {19},
	issn = {1551-3203, 1941-0050},
	doi = {10.1109/TII.2022.3202909},
	language = {en},
	number = {5},
	journal = {IEEE Transactions on Industrial Informatics},
	author = {Huang, Yuxin and Wang, Eric Hao and Liu, Zhaoran and Pan, Licheng and Li, Haozhe and Liu, Xinggao},
	month = may,
	year = {2023},
	pages = {6556--6564},
}

@article{perera_role_2023,
	title = {The role of artificial intelligence-driven soft sensors in advanced sustainable process industries: {A} critical review},
	volume = {121},
	issn = {09521976},
	doi = {10.1016/j.engappai.2023.105988},
	language = {en},
	journal = {Engineering Applications of Artificial Intelligence},
	author = {Perera, Yasith S. and Ratnaweera, D.A.A.C. and Dasanayaka, Chamila H. and Abeykoon, Chamil},
	month = may,
	year = {2023},
	pages = {105988},
}

@article{chai_deep_2022,
	title = {A {Deep} {Probabilistic} {Transfer} {Learning} {Framework} for {Soft} {Sensor} {Modeling} {With} {Missing} {Data}},
	volume = {33},
	issn = {2162-237X, 2162-2388},
	doi = {10.1109/TNNLS.2021.3085869},
	language = {en},
	number = {12},
	journal = {IEEE Transactions on Neural Networks and Learning Systems},
	author = {Chai, Zheng and Zhao, Chunhui and Huang, Biao and Chen, Hongtian},
	month = dec,
	year = {2022},
	pages = {7598--7609},
}

@book{chapelle_semi-supervised_2006,
	title = {Semi-{Supervised} {Learning}},
	isbn = {978-0-262-03358-9},
	publisher = {The MIT Press},
	author = {Chapelle, Olivier and Schölkopf, Bernhard and Zien, Alexander},
	editor = {Chapelle, Olivier and Scholkopf, Bernhard and Zien, Alexander},
	month = sep,
	year = {2006},
	doi = {10.7551/mitpress/9780262033589.001.0001},
}

@article{grimstad2023multi,
  title={Multi-unit soft sensing permits few-shot learning in virtual flow metering},
  author={Grimstad, Bjarne and L{\o}vland, Kristian and Imsland, Lars S},
  journal={arXiv preprint arXiv:2309.15828},
  year={2023}
}

@book{bongers2022causal,
  title={Causal modeling \& dynamical systems: A new perspective on feedback},
  author={Bongers, Stephan R and others},
  year={2022}
}

@inproceedings{falcone2001multiphase,
  title={Multiphase flow metering: current trends and future developments},
  author={Falcone, Gioia and Hewitt, GF and Alimonti, Claudio and Harrison, B},
  booktitle={SPE Annual Technical Conference and Exhibition?},
  pages={SPE--71474},
  year={2001},
  organization={SPE}
}

@manual{corneliussen_2005_handbook,
  title        = {Handbook of Multiphase Flow Metering},
  author       = {Corneliussen, Sidsel and Couput, Jean-Paul and Dahl, Eivind and Dykesteen, Eivind and Frøysa, Kjell-Eivind and Malde, Erik and Moestue, Håkon and Moksnes, Paul Ove and Scheers, Lex and Tunheim, Hallvard},
  organization = {The Norwegian Society for Oil and Gas Measurement},
  month        = {March},
  year         = {2005},
}

\end{document}